\documentclass{article}
\usepackage{ijcai25}

\usepackage{times}
\usepackage{soul}
\usepackage{url}
\usepackage[utf8]{inputenc}
\usepackage[small]{caption}
\usepackage{graphicx}
\usepackage{amsmath}
\usepackage{amsthm}
\usepackage{booktabs}
\usepackage{algorithm}
\usepackage{algpseudocode}
\usepackage{xcolor}
\usepackage{amsthm}
\usepackage{algorithmicx}
\usepackage[switch]{lineno}

\usepackage{balance} 
\usepackage{graphicx}
\usepackage{xspace}
\usepackage{fontawesome5}
\usepackage{tcolorbox}
\tcbuselibrary{breakable}
\usepackage{amsmath}
\usepackage{enumitem} 
\usepackage{stfloats}
\usepackage{float}
\usepackage{subcaption} 
\usepackage{amssymb}

\usepackage{latexsym}

\title{Is Centralized Training with Decentralized Execution Framework Centralized Enough for MARL?}

\author{
Yihe Zhou$^1$
\and
Shunyu Liu$^1$\thanks{Corresponding author.} \and
Yunpeng Qing$^{1}$ \and
Tongya Zheng$^2$\and \\
Kaixuan Chen$^{3,4}$ \and 
Jie Song$^{1}$ \and
Mingli Song$^{3,4}$\\
\affiliations
$^1$Zhejiang University \\
$^2$Zhejiang Provincial Engineering Research Center for Real-Time SmartTech in Urban Security Governance, School of Computer and 
Computing Science, Hangzhou City University\\
$^3$State Key Laboratory of Blockchain and Data Security, Zhejiang University\\
$^4$Hangzhou High-Tech Zone (Binjiang) Institute of Blockchain and Data Security
\emails
\{zhouyihe, liushunyu, qingyunpeng, chenkx, sjie, brooksong\}@zju.edu.cn, doujiang\_zheng@163.com
}

\begin{document}

\maketitle

\begin{abstract}
Centralized Training with Decentralized Execution~(CTDE) has recently emerged as a popular framework for cooperative Multi-Agent Reinforcement Learning~(MARL), where agents can use additional global state information to guide training in a centralized way and make their own decisions only based on decentralized local policies. Despite the encouraging results achieved, CTDE makes an independence assumption on agent policies, which limits agents from adopting global cooperative information from each other during centralized training. Therefore, we argue that the existing CTDE framework cannot fully utilize global information for training, leading to an inefficient joint exploration and perception, which can degrade the final performance. In this paper, we introduce a novel Centralized Advising and Decentralized Pruning~(CADP) framework for MARL, that not only enables an efficacious message exchange among agents during training but also guarantees the independent policies for decentralized execution. 
Firstly, CADP endows agents the explicit communication channel to seek and take advice from different agents for more centralized training. To further ensure the decentralized execution, we propose a smooth model pruning mechanism to progressively constrain the agent communication into a closed one without degradation in agent cooperation capability.
Empirical evaluations on different benchmarks and across various MARL backbones demonstrate that the proposed framework achieves superior performance compared with the state-of-the-art counterparts.  Our code is available at \url{https://github.com/zyh1999/CADP}
\end{abstract}

\section{Introduction\label{intro}}
\begin{figure*}[!t]
  \centering
  \includegraphics[width=0.88\textwidth]{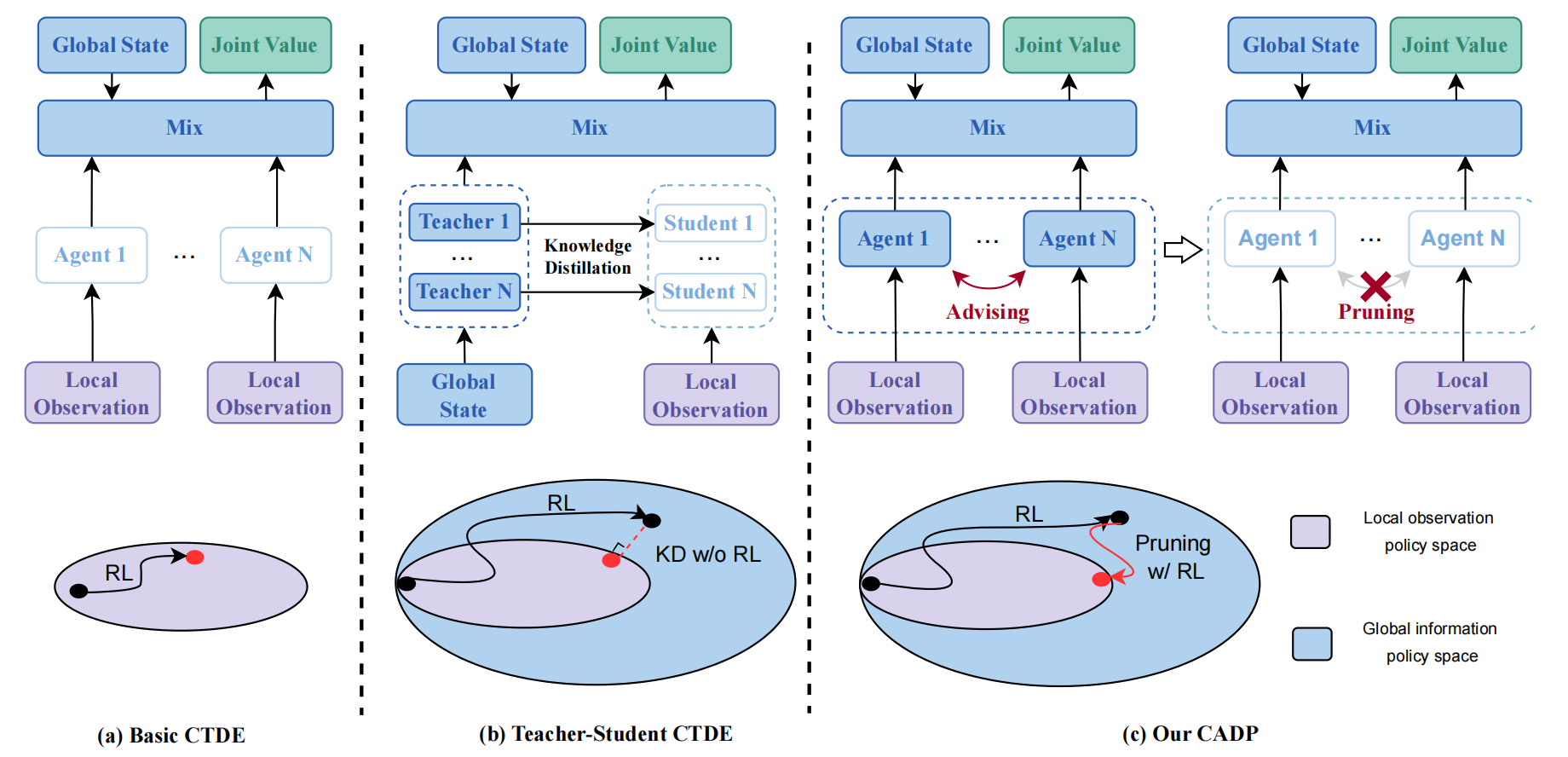}
   \vspace{-0.2cm}
  \caption{Comparisons between existing frameworks and our CADP. (a) Basic CTDE framework. Each agent learns its individual policy by optimizing the joint value of the centralized module with the global state. (b) Teacher-student CTDE framework. This framework introduces knowledge distillation to improve agent learning, where teachers use global information and students  use local information. (c) Our CADP framework. Agents  exchange their advice during centralized training then prune the dependence (still with reinforcement learning)  for decentralized execution. 
  }
  \vspace{-0.2cm}
  \label{fig:illustration}
\end{figure*}

Cooperative Multi-Agent Reinforcement Learning~(MARL) has recently been attracting increasing attention from research communities, attributed to its capability on training autonomous agents to solve many real-world tasks, such as video games~\cite{AlphaStar}, traffic light systems~\cite{wu2020multi} and smart grid control~\cite{power_vipa}. However, learning cooperative policies for various complex multi-agent systems remains a major challenge.  
Firstly, the joint action-observation space grows exponentially with the number of agents, leading to a scalability problem when considering the multi-agent system as a single-agent one to search the optimal joint policy~\cite{QMIX,VDN}. 
Moreover, optimizing agent policies individually also suffers from non-stationarity due to the partial observability constraint~\cite{dagger,iclr2024non-stationary}.
To tackle these problems, Centralized Training with Decentralized Execution (CTDE) is proposed as a popular learning framework for MARL~\cite{MADDPG}. In CTDE, as depicted in Figure~\ref{fig:illustration}(a), decentralized agent policies are trained by a centralized module with additional global state information, while agents select actions only based on their own local observation without any communication

In recent years, the CTDE framework has been widely used in MARL, including Value Decomposition~(VD) methods~\cite{VDN,QMIX,QPLEX,QTRAN,WQMIX,CIA,liu2024OPT,kapoor2024} and Policy Gradient~(PG) methods~\cite{MADDPG,COMA,MAPPO,HATRPO}, which achieves the state-of-the-art performance in different benchmarks.
Despite its promising success, we argue that the centralized training in CTDE is not centralized enough. This is to say, the existing CTDE framework cannot take full advantage of global information for centralized training. Specifically, agent policies are assumed to be independent of each other~\cite{MACPF}, and the existing CTDE framework only introduces global information in the centralized module, while agents are not granted access to global information even when centralized training. This partial observability limits agents to search better global joint policy~\cite{dagger,PTDE,CTDS}.

To remedy this issue, several prior efforts propose to design a teacher-student CTDE framework~\cite{dagger,PTDE,CTDS}, as depicted in Figure~\ref{fig:illustration}(b). These works enable teacher agents to use the global state information during centralized training, while student agents with local observation can imitate the behaviors from teacher agents via knowledge distillation. However, these works just simply take the additional state information as the input of agent policy, which follows the independence assumption on agent policies. Therefore, the agents still make their own decisions without considering the policies of other agents during centralized training. In this way, the expressiveness of the joint policy is inevitably limited, leading to an inefficient joint exploration and perception, which can degrade the final performance.

In this paper, we propose a novel Centralized Advising and Decentralized Pruning framework, termed as CADP, to enhance basic CTDE with global cooperative information. As depicted in Figure~\ref{fig:illustration}(c), CADP enables agents to exchange advice with each other instead of only using global state information during centralized training. This approach is aligned with a common way of human communication, where humans often offer helpful and tailored advice based on their knowledge and beliefs instead of simply providing their own information~\cite{stenning2006introduction}. Therefore, the centralized advising mechanism allows agents to deliver team-oriented actions for better cooperation.
To generate the final decentralized policies, we further propose to smoothly prune the dependence relationship among agents via a dedicated auxiliary loss function.

Our main contribution is the dedicated attempt that adopts agent communication to enhance basic CTDE framework for fully centralized training with decentralized execution. We propose a novel Centralized Advising and Decentralized Pruning~(CADP) framework to promote explicit agent cooperation during training while still ensure the independent policies for execution.  CADP is designed to provide a new general training framework for different MARL methods based on CTDE. Experiments conducted on various benchmarks show that the proposed CADP framework yields results superior to the state-of-the-art  methods.


\section{Related Works} \label{related}

\textbf{Teacher-student CTDE Framework.}
Conventional  CTDE-based methods fall short of fully utilizing global information for training, leading to an inefficient exploration and perception of the joint-policy and resulting in performance degration. Therefore, several recent works attempt to improve CTDE with the teacher-student framework.
IGM-DA~\cite{dagger} firstly proposes centralized teacher with decentralized student frameworks, where the teacher models use the global state while the student models use the partial observation. It adds an additional knowledge distillation loss  to enable the teacher model to assist in training the student model. In CTDS~\cite{CTDS}, the teacher model still uses the agent observation, but the field of view is set to infinite during centralized training. PTDE~\cite{PTDE} trains a  network to aggregate local observation and global state, resulting in a better representation of agent-specific global information.
Most of these methods focus on distillation to transfer the knowledge from the teacher to the student for decentralized execution, which is essentially an offline imitation learning. Thus, the agents still ignore the policies of other agents and make their own decision during centralized training, resulting in an inefficient cooperative exploration.

\textbf{Communication in MARL.}
Communication~\cite{Bottleneck,CommNet,li2022ace,nayak2023scalable,iclr2024contrastive} is also widely used in MARL, which facilitates information transmission between agents for effective collaboration when agents can only access their own observations. Recently, many methods have  used attention as the main mechanism of communication~\cite{das2019tarmac,ATOC,MAAC,MAIC,DAACMP,iclr2024graph}, where self-attention can be considered as a message exchanging mechanism with other agents. The communication paradigm allows agents to communicate during both the training and execution stages. In contrast, CTDE emphasizes centralized training of agents while they execute their individual policies in a decentralized no-communication manner. Furthermore, to address communication constraints, various message pruning methods~\cite{IMAC,GACML,NDQ,MAIC,I2C} propose 
to compress and refine communication information, choose with whom to communicate. Furthermore, MACPF~\cite{MACPF} offers a method involving information transmission during training without necessitating it during execution. However, it only enables unidirectional forwarding of previous agents' actions,   which does not qualify as comprehensive  mutual communication.

\section{Preliminary}
\subsection{Dec-POMDP}

We consider a fully cooperative multi-agent task as the \textit{Decentralized Partially Observable Markov Decision Process}~(\text{Dec-POMDP}),
which is defined as a tuple $\langle \mathcal{A}, \mathcal{S}, \mathcal{U}, P, r, \Omega, O, \gamma \rangle$, where $\mathcal{A} = \{a_n\}_{n=1}^N$ is the set of $N$ agents and $s\in\mathcal{S}$ is the global state of the environment. At each time step $t$, each agent $a_n \in \mathcal{A}$ receives an individual partial observation $o^n_t \in \Omega$ drawn from the observation function $O(s_t, a_n)$, then each agent chooses an action $u^n_t \in \mathcal{U}$ which forms joint action $\boldsymbol{u}_t \in \mathcal{U}^N$. This causes a transition to the next state $s_{t+1}$ according to the state transition function $P(s_{t+1} | s_t, \boldsymbol{u}_t):\mathcal{S}\times\mathcal{U}^N\times\mathcal{S} \to [0,1]$. The reward function which is modeled as  $r(s_t, \boldsymbol{u}_t):\mathcal{S}\times\mathcal{U}^N\to \mathbb{R}$ and $\gamma \in [0, 1)$ is the discount factor. Each agent $a_n$ has an action-observation history $\tau^n \in \mathcal{T} \equiv (\Omega \times \mathcal{U})^*$, on which it conditions a
stochastic policy $\pi^n(u^n|\tau^n):\mathcal{T}\times\mathcal{U}\to[0,1]$. The joint action-observation history is defined as $\boldsymbol{\tau}\in \mathcal{T}^N$. In this work, the joint policy $\boldsymbol{\pi}$ is based on joint action-value function $Q^{tot}_t(s_t,\boldsymbol{u}_t)=\mathbb{E}_{s_{t+1:\infty},\boldsymbol{u}_{t+1:\infty}}{\left[\sum_{i=0}^{\infty}{\gamma^i r_{t+i}} \mid s_t,\boldsymbol{u}_t\right]}$ . The final goal is to get the optimal policy $\boldsymbol{\pi}^*$ that maximizes the joint action value.

\subsection{Value Decomposition in MARL}
Value Decomposition~(VD) is a useful technique  in cooperative MARL to achieve effective Q-learning~\cite{VDN,QMIX,QPLEX,sharpley_credit,jiang2021action,A2PO_yunpeng}. 
It aims to learn a joint action-value function $Q^{tot}$ to estimate the expected return given current global state $s_t$ and joint action  $\boldsymbol{u}_t$. 
To realize VD, a mix network $f(\cdot ;\theta_{\upsilon})$ with parameters $\theta_{\upsilon}$ is adopted as an approximator to estimate the joint action-value function $Q^{tot}$. 
$f(\cdot ;\theta_{\upsilon})$ is introduced to merge all individual values into a joint one $Q^{tot} = f(\boldsymbol{q};\theta_{\upsilon })$, where $\boldsymbol{q} = [Q^n]_{n=1}^N \in \mathbb{R}^N$ and $Q^n$ with shared parameters $\theta_{a}$ is the action-value network of each agent $a_n$. Usually, $f(\cdot ;\theta_{\upsilon})$ is enforced to satisfy the Individual-Global-Max~\cite{QTRAN} principle.
Therefore, the optimal joint action can be easily derived by
independently choosing a local optimal action from each local Q-function $Q^n$, which enables Centralized Training and Decentralized Execution~(CTDE). The learnable parameter $\theta=\{\theta_{a},\theta_{\upsilon }\}$ can be updated by minimizing the Temporal-Difference~(TD) loss as:
\begin{align}
    \mathcal{L}_{TD}(\theta) =  \mathbb{E}_{\mathcal{D}} \left[\left(y^{tot} - Q^{tot}\right)^2\right]. \label{TDloss}
\end{align}
where $\mathbb{E}[\cdot]$ denotes the expectation function, $\mathcal{D}$ is the replay buffer of the transitions, $y^{tot}=r+\gamma \hat{Q}^{tot}$ is the one-step target and $\hat{Q}^{tot}$ is the target network~\cite{DQN15}. Additionally, owing to the fact that the partial observability often limits the agent in the acquisition of information, the agent policy usually uses past observations from history~\cite{VDN}.


\subsection{Policy Gradient in MARL}
Policy gradient (PG)~\cite{MAPPO,MADDPG,COMA} has been proposed as a competent alternative to  directly optimize the policy. In the domain of cooperative MARL, the PG mechanism complies with the CTDE constraint through the learning of an individual actor $\pi^n(u^n|\tau^n)$, and a centralized critic $V(s): \mathcal{S} \to \mathbb{R}$, for all agents. To leverage global information during centralized training, the value functions $V^n$ typically incorporate the global state $s$ as input to ensure accurate estimation of the expected value.
By adopting this decentralized execution approach, the consequent implicit joint policy achieves a fully independent factorization and agent policies are assumed to be independent of each other~\cite{fu2022revisiting}:
$
 \boldsymbol{\pi}(\boldsymbol{u} \mid {\boldsymbol{\tau}}) = \prod_{n=1}^N \pi^{n}\left(u^n \mid \tau^n\right).
$
PG directly maximizes the expected discounted return $R_t=\sum_{i=0}^{\infty} \gamma^{i} r_{t+i}$ as the objective. Thus, the loss function is defined as $\mathcal{L}_{actor}=-\mathbb{E}_{\boldsymbol{\pi}} [R_t]$.
To optimize the actor $\pi$, we can perform policy gradient~\cite{sutton2018reinforcement} as:
\begin{align} \label{ACloss}
\nabla_{\theta_{\boldsymbol{\pi}}} \mathcal{L}_{AC}(\theta_{\boldsymbol{\pi}})=-\mathbb{E}_{s \sim p^{\boldsymbol{\pi}}, \boldsymbol{u} \sim \boldsymbol{\pi}} 
\left[ \sum_{t=0}^T R_t \nabla_{\theta_{\boldsymbol{\pi}}} \log \boldsymbol{\pi} \left(\boldsymbol{u_t} \mid \boldsymbol{\tau_t}\right)\right],
\end{align}
where $p^{\boldsymbol{\pi}}$ is the state distribution. In particular, $R_t$ is often replaced by $r_t+\gamma V(s_{t+1})-V(s_t)$ which is calculated by the critic function to reduce the high variance.

\section{Method}

\begin{figure*}[!t]
  \centering
  \includegraphics[width=0.88\textwidth]{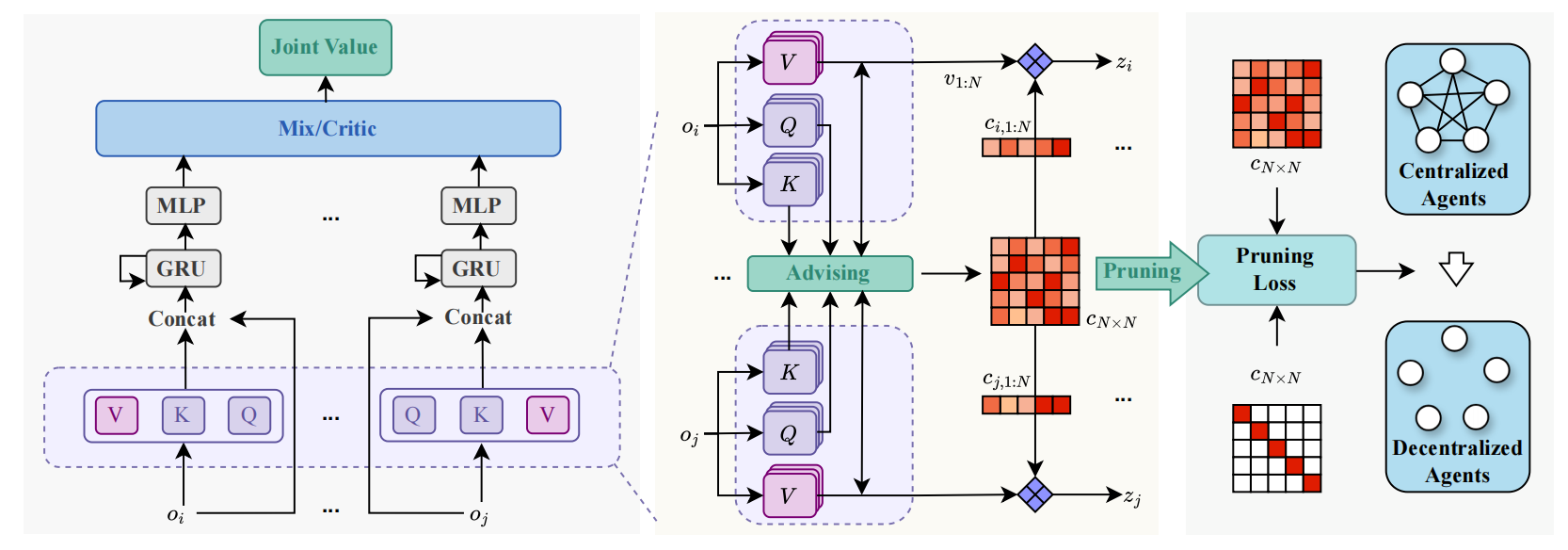}
   \caption{Illustrative diagram of the proposed Centralized Advising and Decentralized Pruning~(CADP) framework. At  centralized training stage, the agent model will use the $Q$, $K$, $V$ modules, while at  decentralized execution stage, the agent model only uses $V$ module.}
  \label{fig:method}
  \vspace{-0.3cm}
\end{figure*}

To introduce global cooperative information for agent training, we propose the Centralized Advising and Decentralized Pruning~(CADP) framework, as shown in Figure~\ref{fig:method}.
In general, CADP performs CTDE to enable each agent to learn its individual policy network.
At the heart of our design is introducing the explicit cooperative information exchanging of agents for sufficient centralized training to enhance the agent policy network.
To cope with the decentralized execution paradigm, we design a model self-pruning mechanism, which prompts the centralized model to evolve into a decentralized model smoothly. The overall framework is finally summarized. Besides, for the sake of clarity, we employs the VD method as an illustrative example to introduce the proposed CADP framework, while CADP is also readily applicable to the PG method.


\subsection{Advice Exchanging}\label{exchanging}

The widely adopted CTDE framework only introduces the global state  for agents in the mix/critic module, leading to that an agent policy network only perceives its local observation instead of the global states. In contrast, we design a novel centralized training scheme to augment the agent policy network from the local information of an individual agent to the global cooperative information from all agents, inferring better action decisions.

Formally, we employ an agent's confidence $c$ for all agents to highlight its personalized confidence weights of other agents when receiving interchangeable cooperative advice from them, where the higher confidence corresponds to the more useful information of agents.
The whole process can be reduced to a self-attention mechanism~\cite{attention}, where we set messages key $k$ and value $v$, respectively, while confidence $c$ is considered as the dot product of the key $k$ of other agents and  query $q$ of itself. 
$q$, $k$ and $v$ are all linear projections of the local observation $o$. The formula is written as:
\begin{align}
    & \alpha_{i,j} :=  \frac{q_i \cdot k_j} {\sqrt{d_x}},  \\ 
    & c_{i,j} := \frac{exp(\alpha_{i,j})}{\sum_{k=1}^{N} exp(\alpha_{i,k})},
\end{align}
where $k_j$ means the message key $k$ of agent $j$ and $q_i$ means the query $q$ of agent $i$, $d_x$ is the scaling coefficient and $c_{i,j}$ is the confidence from agent $i$ to agent $j$.
Finally, the aggregating information $z_i$ of the agent $i$ is obtained by taking the weighted sum of the value according to the confidence $c_{i,1:N}:= \left( c_{i,1}, c_{i,2},\cdots, c_{i,N} \right)$:
\begin{align}
    z_{i} := \sum_{j = 1}^{N}c_{i,j} \cdot  v_j, 
\end{align}
where $v_{j}$ means the value $v$ of agent $j$. Through this step, each agent  refers to the cooperative information of others. Then, we combine the aggregating information $z$ in the previous step with the agent's own local information $h$, and finally output the action value $Q$: 
\begin{align} 
    & h_i^t := GRU([z_i,o_i],h_i^{t-1}),\quad Q^i :=  MLP(h_i^t),\label{Qi}
\end{align}
where $GRU(\cdot,\cdot)$ stands for Gated Recurrent Unit. Notably, we have incorporated residual connections into the input of the GRU network. This short-circuit mechanism allows us to simultaneously leverage representations with and without advice exchanging, for enhancing training stability and performance.

Since our work focuses on the agent policy module, we can adopt different mix modules such as VDN~\cite{VDN}, QMIX~\cite{QMIX} and QPLEX~\cite{QPLEX} to generate $Q^{tot}$. Besides, if the final output of our agent module is a policy distribution $\pi^i$ instead of $Q^i$ (performing  normalization after Equation~\ref{Qi}), we can also employ MAPPO~\cite{MAPPO}.

\subsection{Model Self-Pruning}
In pursuit of facilitating decentralized execution, the current model needs to evolve into the decentralized model depending only on itself, rather than global information. In this step, we design a simple yet effective model self-pruning method to achieve this.
We claim that if the following conditions are satisfied, the agent model is a decentralized model:
\begin{align}
      c_{i,1:N} = \mathbf{e}^i, \forall i \in [1,N],
\end{align}
where $\mathbf{e}^i$ means the $i$-th standard basis vector (an one-hot vector). In this way, we can directly apply the agent model to decentralized execution as there is no advice from other agents used. Furthermore, we just apply the value $v$ to produce the output $z$ without key $k$ and other components. For the convenience of expression, we refer to the model that only requires their own values $v$ without the self-attention mechanism as the decentralized model (CADP (D)). On the other hand, the model using self-attention mechanism is referred to as the centralized model (CADP (C)).
A decentralized model actually presents that the agent's confidence $c_{i,1:N}$ of all agents is equal to the one-hot vector $\mathbf{e}^i$ in the execution stage.
It is required to smoothly swap from the centralized training with exchanging confidence to the decentralized execution with the one-hot confidence $c_{i,1:N}$.
Therefore, we design an auxiliary loss function named pruning loss $\mathcal{L}_p$ to help the decentralized agent gradually alleviate the dependence of other agents, which is given as:
\begin{align}
   & \mathcal{L}_p(\theta_a)=\sum_{i=1}^{N}D_{KL}(\mathbf{e}^i \|c_{i,1:N}),
\end{align}
where $\theta_a$ means  the parameters in the agent policy module, and $D_{KL}$ means  the Kullback Leibler (KL) Divergence. In the pruning loss, smaller  $\mathcal{L}_p$ means that agents rely less on the other agents.

\subsection{Overall Framework}
In our framework, we  adopt advice exchanging step  at agent policy module to produce more thoughtful and team-oriented action decisions. After the model achieves satisfactory performance, we start to use the  pruning loss  $\mathcal{L}_p$,  prompting the centralized model to evolve into a decentralized one smoothly.
Our mix module implementation uses QMIX~\cite{QMIX} as
a basic backbone for its robust performance and its simplicity of architecture, but it is readily applicable to the other mix/critic method since our framework focuses on the agent policy module. 

To sum up, training our  CADP framework contains two main loss functions. 
The first one is naturally the original TD
loss $\mathcal{L}_{TD}$ mentioned on Equation~ \ref{TDloss}, which enables each agent to learn its individual
agent policy by optimizing the joint-action value of the mix module. Unlike  IGM-DA~\cite{dagger} and CTDS~\cite{CTDS},
to avoid performance degradation and reduce computing costs, our pruning loss  $\mathcal{L}_p$  is not introduced at the very beginning. We add pruning loss when the centralized model reaches a high level. Therefore, the total loss of our framework is formulated as follows. 
(Here we take the method of value decomposition as an example. For the PG method, we just need to add the pruning loss to the loss of actor $\mathcal{L}_{AC}$ mentioned on Equation~ \ref{ACloss}):
\begin{align}
\mathcal{L}_{tot}(\theta) =  \mathcal{L}_{TD}(\theta_{mix},\theta_a) + \sigma(t) \mathcal{L}_{p}(\theta_a),
\end{align}
where $\theta_{mix}$ stands for the parameters of mix network in our method and $t$ is the timestep of the training. $\sigma(\cdot)$ is a threshold function which is defined as follows:
\begin{align}
 \sigma(t) = \left\{\begin{array}{ll}
\alpha &  if \quad t \geq T, \\
0 & Otherwise,
\end{array} \right.
\end{align}
where $T$ is a hyperparameter and $\alpha$ is the coefficient for trading off loss term. From a practical application perspective, it suffices to incorporate the pruning loss after the centralized model  achieves satisfactory performance in online learning.  Nevertheless, in Section ~\ref{ablation}, we still conduct ablation experiments for $T$ to further investigate its impact. In addition, we provide pseudocode  in Appendix D.

\section{Experiments}\label{exp}
To demonstrate the effectiveness of the proposed  CADP framework, we conduct experiments on the StarCraft II micromanagement challenge and Google Research Football benchmark. 
We aim to answer the following questions: (1) Can CADP outperforms the methods under traditional CTDE framework? (Section \ref{traditional_CTDE}) (2) Can CADP outperforms the methods under teacher-student CTDE framework? (Section \ref{teacher-student_CTDE}) (3) Can CADP perform better than the message (communication) pruning methods under the CTDE constraints? (Section \ref{pruning}) (4) Can CADP, as a universal framework, be applied to different methods and achieve improvement? (Section \ref{plug}) (5) In comparison to traditional CTDE methods, does CADP require excessive time overhead? (Appendix C).
Besides, visualization is given in Appendix E.

Based on these questions, we  opt to compare three major categories of baseline methods: traditional CTDE-based methods, teacher-student CTDE framework methods and the message pruning methods under the CTDE constraints. Each category will be compared with CADP.  Our CADP framework uses QMIX~\cite{QMIX} as a basic backbone for its robustness and its simplicity of architecture, but it is readily applicable to the other mix/critic backbone. We show it at Section \ref{plug}. The detailed hyperparameters are given in Appendix B , where the common  hyperparameters across different methods are consistent for comparability. We also provide additional expereiments in  Appendix B.

\subsection{Comparison with Traditional CTDE}\label{traditional_CTDE}
\begin{figure*}[!t]
    \centering
    \subfloat{\quad\quad\includegraphics[width=0.66\textwidth]{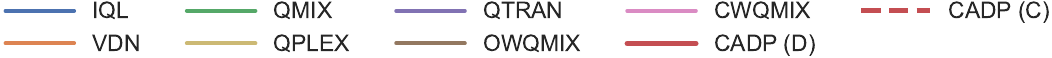}}\vspace{-0.1em}\\    
    \addtocounter{subfigure}{-1}
    \subfloat[5m\_vs\_6m~(Hard)]{\includegraphics[width=0.30\textwidth]{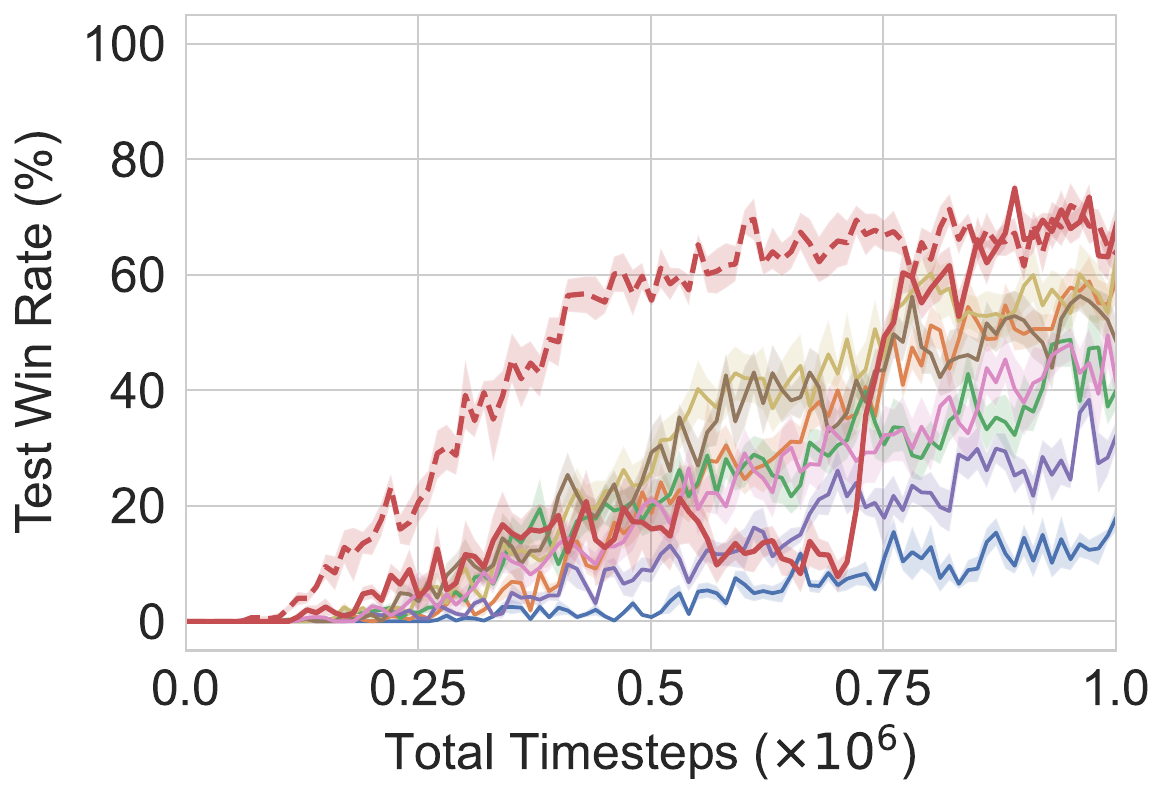}}
    \subfloat[corridor~(Super Hard)]{\includegraphics[width=0.30\textwidth]{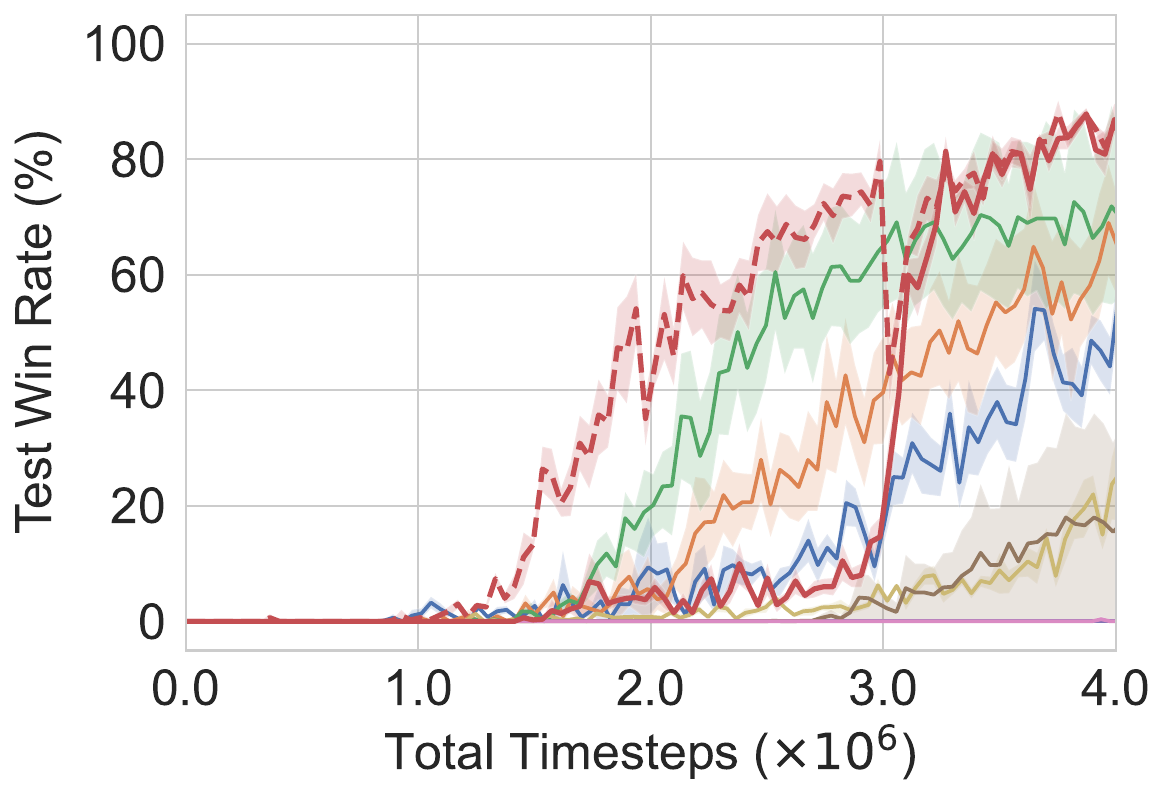}}
    \subfloat[3s5z\_vs\_3s6z~(Super Hard)]{\includegraphics[width=0.30\textwidth]{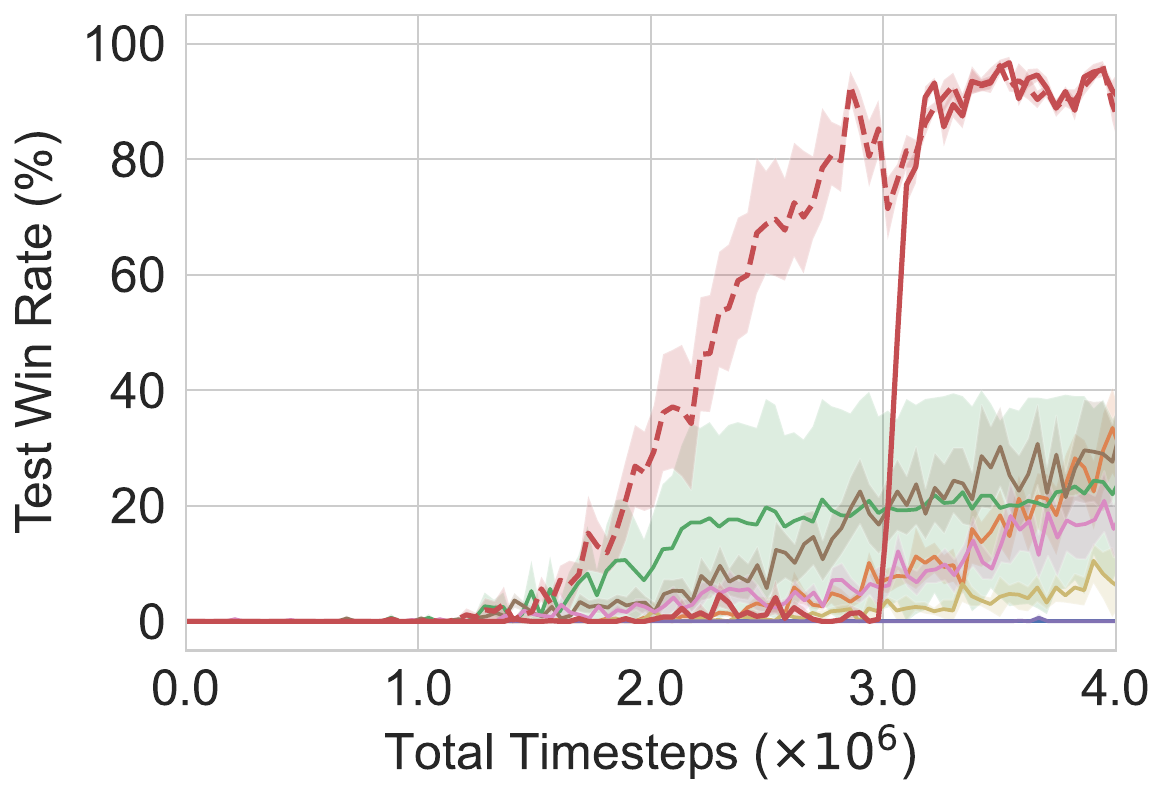}} \\
    \vspace{0.1em}
    \subfloat{\quad\;\includegraphics[width=0.50\textwidth]{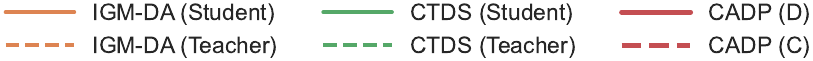}}\vspace{-0.1em}\\  
     \addtocounter{subfigure}{-1}
    \subfloat[5m\_vs\_6m~(Hard)]{\includegraphics[width=0.30\textwidth]{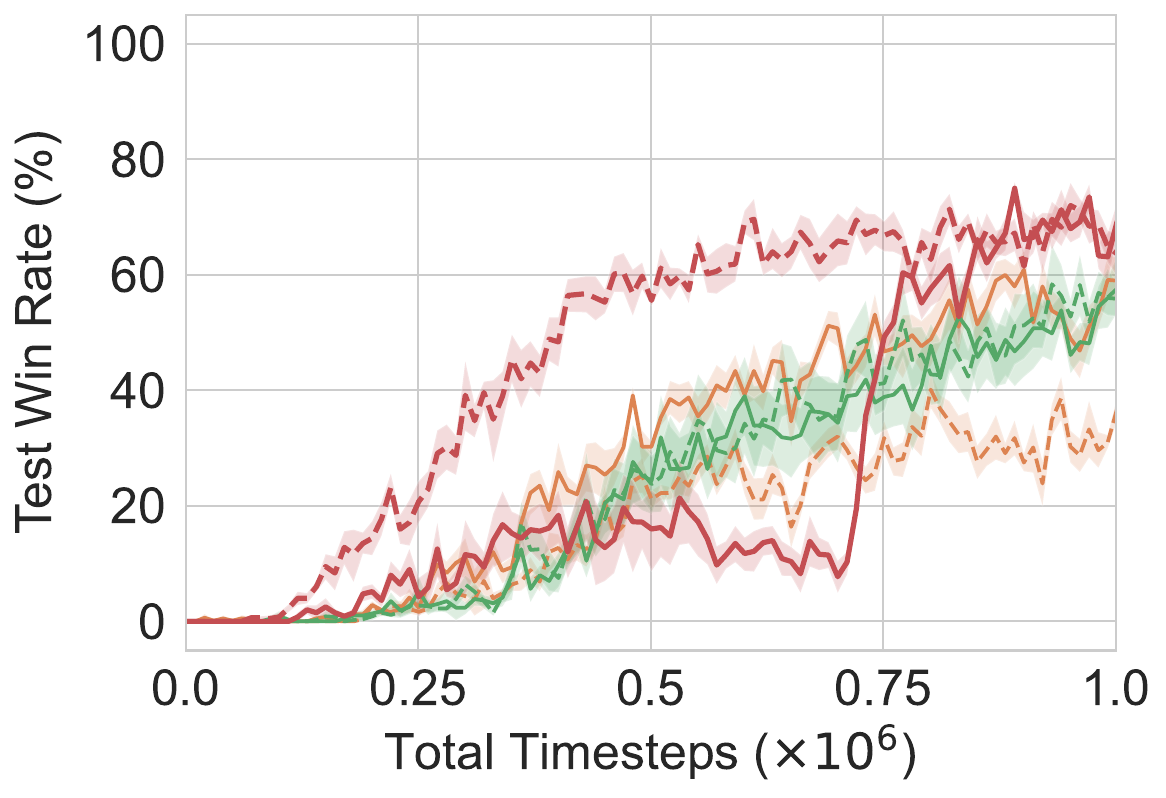}}
    \subfloat[corridor~(Super Hard)]{\includegraphics[width=0.30\textwidth]{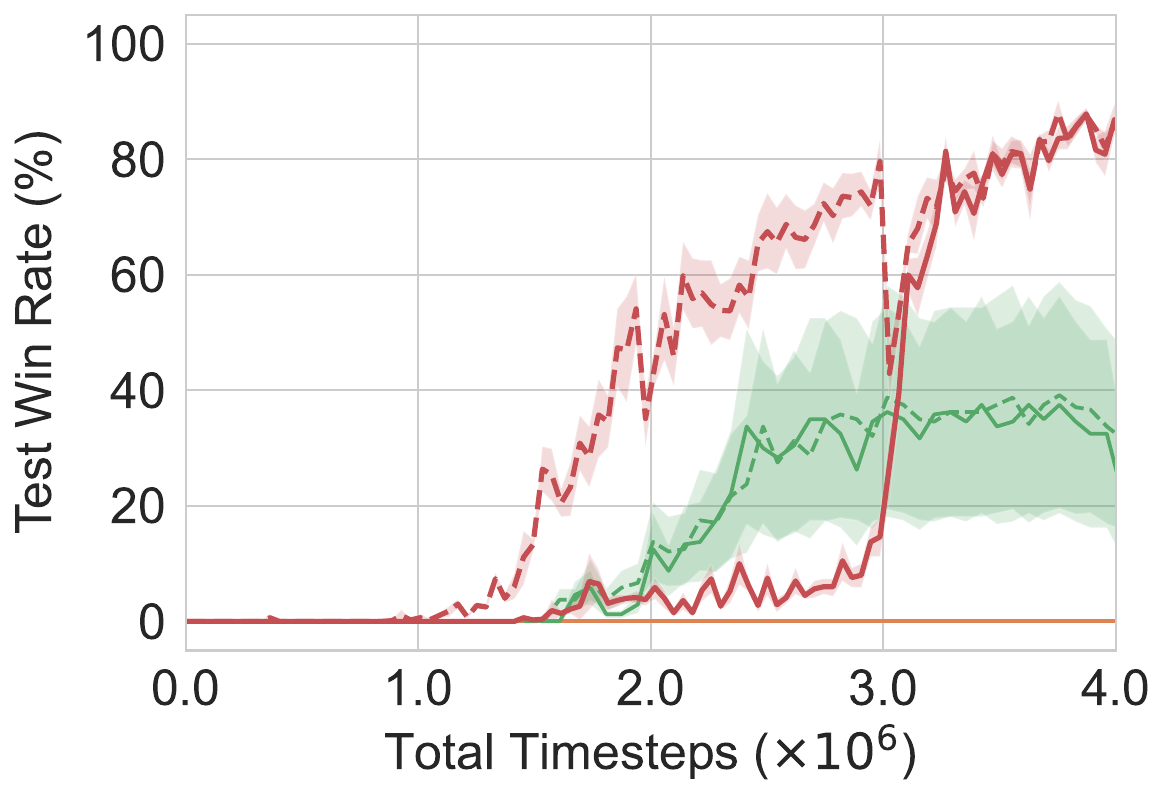}}
    \subfloat[3s5z\_vs\_3s6z~(Super Hard)]{\includegraphics[width=0.30\textwidth]{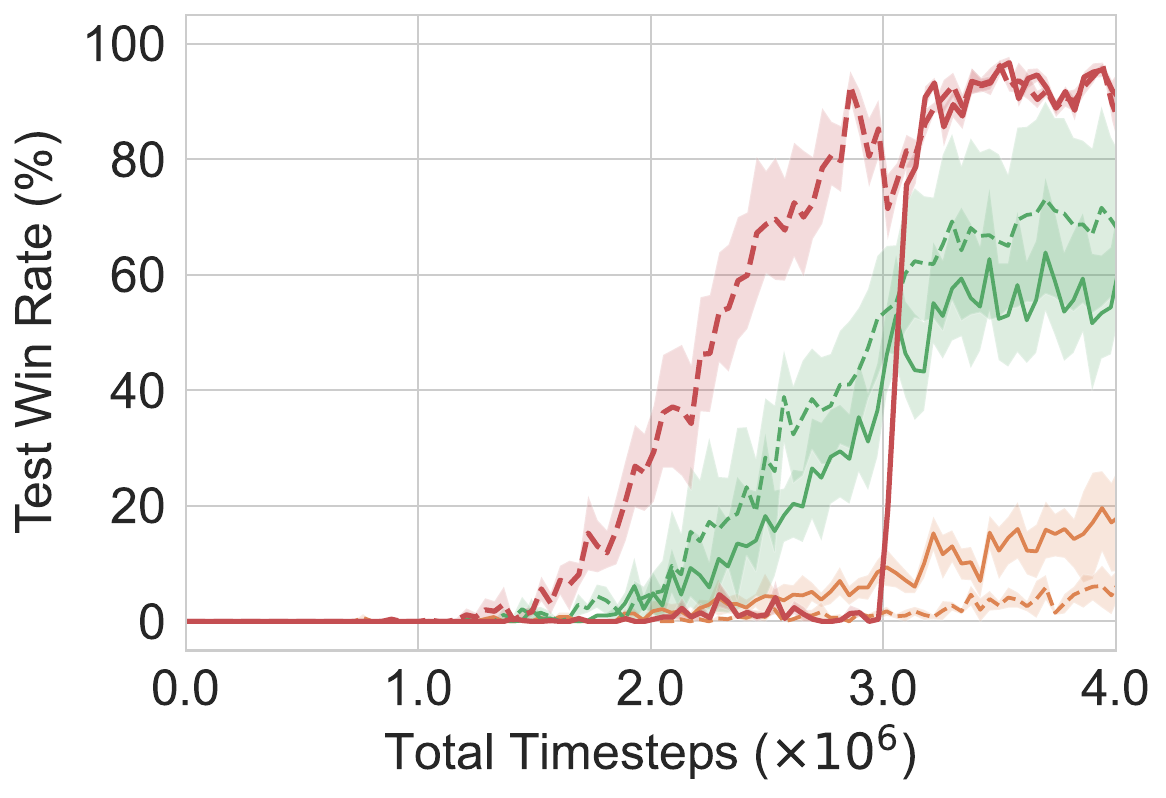}}
    \caption{Learning curves of our method and baselines on the SMAC scenarios. 
    (Upper) Comparision with the methods under the CTDE and DTDE frameworks. (Lower)  Comparision with the methods under the teacher-student CTDE framework.
     CADP(C) means our centralized model, while  CADP(D) means our decentralized model which will be used for decentralized execution.}
    \label{fig:results}
     \vspace{-0.1cm}
  \end{figure*}

We select several traditional CTDE methods: VDN~\cite{VDN}, 
 QMIX~\cite{QMIX}, QTRAN~\cite{QTRAN}, QPLEX~\cite{QPLEX}, 
 CWQMIX~\cite{WQMIX}, OWQMIX ~\cite{WQMIX} and one DTDE-based methods 
 (regarded as a special case of CTDE): IQL~\cite{iql}  for comparison.
The experimental results on different scenarios are shown
in Figure~\ref{fig:results}. It can be seen that our proposed method successfully improves the final performance in the challenging tasks. Especially in the most difficult homogeneous scenario~(\emph{3s5z\_vs\_3s6z}) due to the different unit types, the large number of entities and the great disparity in strength between the two teams, our method can outperform baselines by a large margin.

\subsection{Comparison with Teacher-Student CTDE}\label{teacher-student_CTDE}
We also  compare with the teacher-student CTDE framework methods: CTDS~\cite{CTDS} and IGM-DA~\cite{dagger}. The experimental results on different scenarios are shown
in Figure~\ref{fig:results}. Similarly to the  previous section, our method has better final performance in the challenging tasks, not only compared with the methods under teacher-student CTDE framework. Especially in the super hard scenarios~(\emph{3s5z\_vs\_3s6z}) and ~(\emph{corridor}) due to the different unit types or the large number of entities and the great disparity in strength between the two teams, our method outperforms baselines by a large margin. 
In addition, after adding the pruning loss, the performance of our decentralized model rises dramatically from almost zero to near the performance of our centralized model. By contrast, in  methods teacher-student CTDE framwork, the performances of student models and teacher models  climb together  slowly and even sometimes the teacher models fail for training. Furthermore, our CADP method is the only  method which can ultimately reach almost consistent performance for decentralized model and centralized model in all scenarios.

  \begin{figure}[!t]
    \centering
    \subfloat{\;\includegraphics[width=0.43\textwidth]{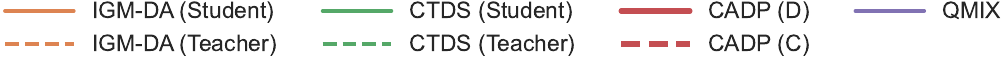}}
    \\
     \addtocounter{subfigure}{-1}
    \subfloat[3\_vs\_1\_with\_keeper]{\includegraphics[width=0.23\textwidth]{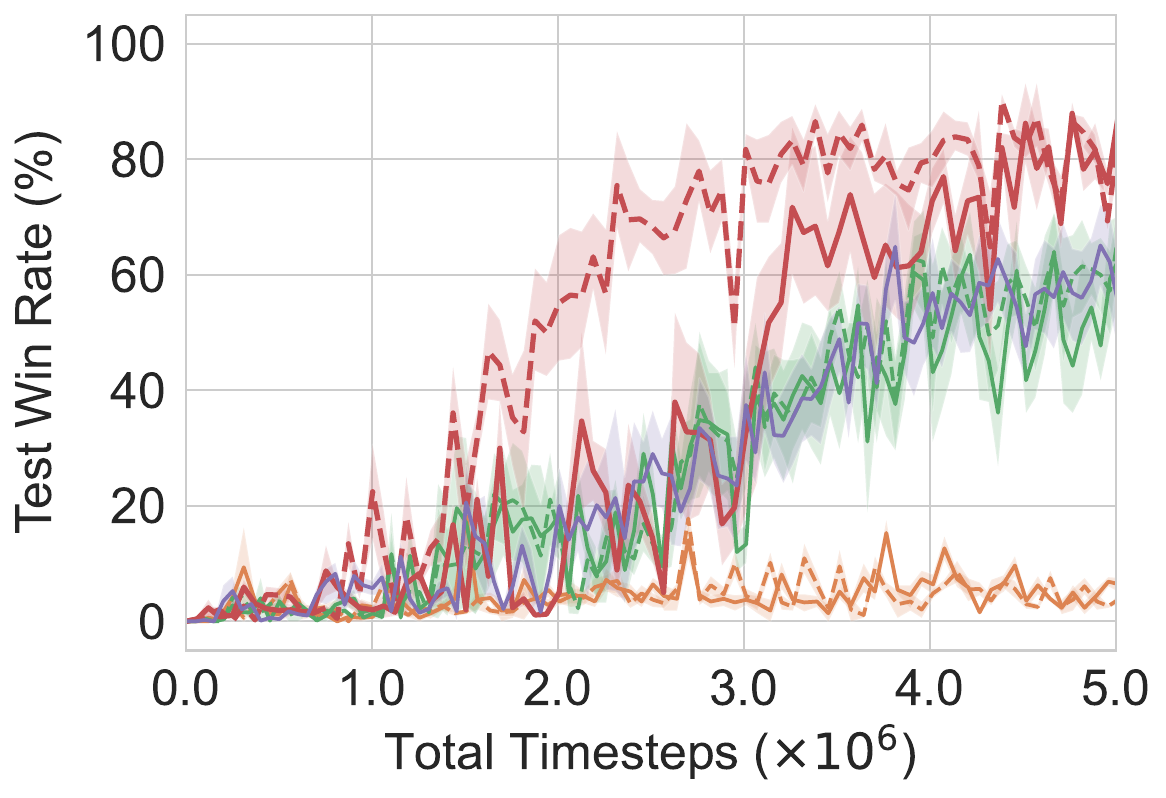}}
    \subfloat[counterattack\_easy]{\includegraphics[width=0.23\textwidth]{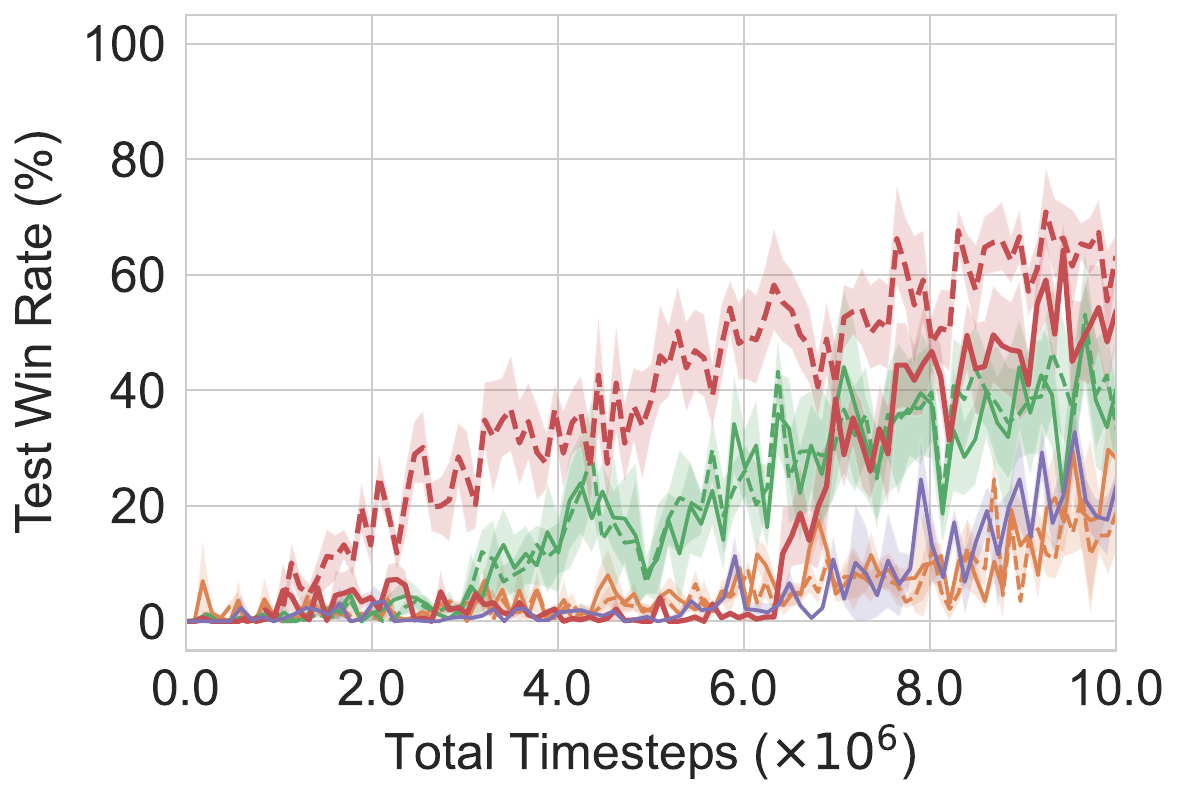}}
    \caption{Learning curves of our CADP method and baselines on the  Google Research Football (GRF) scenarios.}
    \label{fig:grf}
     \vspace{-0.5cm}
  \end{figure}

Besides, we test teacher-student CTDE framework methods on the Google Research Football (GRF) \cite{GRF}  benchmark. Unlike StarCraft II, individual observations in the Google Research Football~(GRF)  are not partial observation, which contains as much information as global state. In this case, the teacher-student CTDE framework will no longer have a significant advantage over  methods under traditional CTDE framework theoretically, since the teacher model does not have more input information than the student model and has no ability to instruct the student model.
The experimental results on different scenarios are shown
in Figure~\ref{fig:grf}. It can be seen that only our method  significantly outperforms QMIX~\cite{QMIX}. In addition, in the method of the teacher-student framework, the student model  always maintains almost the same performance as the teacher model, which  verifies that the teacher model does not have the ability to guide the student model in this case and also shows that using global cooperative information is more powerful than simply providing more observation information. In GRF benchmark, we set $T=3M$ in \emph{3\_vs\_1\_with\_keeper} scenario and $T=6M$ in \emph{counterattack\_easy} scenario respectively. We can see an obvious  improvement of our method after adding the pruning loss $\mathcal{L}_p$ at timestep $T$ in both.


\subsection{Comparison with Message  (Communication) Pruning Methods under CTDE Constraint}\label{pruning}
 Furthermore, we  test some message pruning methods, GACML~\cite{GACML}, NDQ~\cite{NDQ},  MAIC~\cite{MAIC} and MACPF~\cite{MACPF} under CTDE constraint. GACML~\cite{GACML} and MAIC~\cite{MAIC} are attention-based methods, while the other is not. During training, we allow them to communicate and apply message pruning, but during execution, we cut off all communication. To ensure a fair comparison, we also used QMIX~\cite{QMIX} as the backbone for these  methods. The experimental results in the table~\ref{tab:communication} show that these methods are almost not better than basic QMIX~\cite{QMIX} which means when communication is completely cut off, the  performance of these methods is not good.

\begin{table}[!t]
  \centering
  \small
  \caption{The comparison of test win rate between our CADP method and the message pruning methods.}
   {%
    \begin{tabular}{cccc}
      \toprule
      \multicolumn{1}{c}{\textbf{Method}} & \textbf{5m\_vs\_6m} & \textbf{corridor} & \textbf{3s5z\_vs\_3s6z} \\
      \midrule
      {GACML} & 0.46 $\pm$ 0.10 & 0.30 $\pm$ 0.33 &  0.32 $\pm$ 0.36 \\ \specialrule{0em}{1pt}{1pt}
      {NDQ}  & 0.38 $\pm$ 0.09 &  0.42 $\pm$ 0.21 &  0.00 $\pm$ 0.00 \\  \specialrule{0em}{1pt}{1pt}
      {MAIC}  & 0.28 $\pm$ 0.10 &  0.01 $\pm$ 0.01 &  0.32 $\pm$ 0.45 \\  \specialrule{0em}{1pt}{1pt}
      {MACPF}  & 0.20 $\pm$ 0.18 &  0.67 $\pm$ 0.39 &  0.16 $\pm$ 0.14 \\  \specialrule{0em}{1pt}{1pt}
      \midrule
      {QMIX (CTDE)} &  0.43$\pm$ 0.13 & 0.70 $\pm$ 0.35 & 0.24 $\pm$ 0.36\ \\ \specialrule{0em}{1pt}{1pt}
      {QMIX (CADP)} & \textbf{0.68$\pm$ 0.08} & \textbf{0.84 $\pm$ 0.03} & \textbf{0.93 $\pm$ 0.03} \\
      \bottomrule
    \end{tabular}%
    }
  \label{tab:communication}%

\end{table}%

These experimental results indicate that although message pruning can reduce communication, it is not suitable for CTDE scenarios where there is no communication at all. These message pruning methods, except MACPF~\cite{MACPF}   focus on compressing and refining communication information or selecting who to communicate with, which results in smaller bandwidth. As for MACPF~\cite{MACPF}, the poor performance can mostly be attributed to the insufficient information exchanging, which is solely the one-hot vector representing the actions of the previous agents. In contrast,  our CADP framework is designed to enhance the CTDE framework, which does not allow any communication at all during the decentralized execution phase. Our CADP framework first trains a well-performing communication model, and then gradually and dynamically distills its knowledge into a non-communication model. 


\subsection{Ablation Study} \label{ablation}

\textbf{Different backbones.} \label{plug}
To further verify the generability of our CADP framework, we test value-based methods: VDN~\cite{VDN}, QMIX~\cite{QMIX}, QPLEX~\cite{QPLEX} and policy-based method: MAPPO~\cite{MAPPO} under the CADP framework. Experimental results in Table~\ref{tab:example} indicate that all these methods have shown an improvement when using the CADP framework.
Especially in the most difficult homogeneous scenario (\emph{3s5z}\_\emph{vs}\_ \emph{3s6z}), methods with our CADP framework outperform baselines by a large margin. This observation  indicates the versatility and effectiveness of our proposed CADP framework in catering to various CTDE-based methods, particularly in scenarios with high levels of difficulty.

\begin{table}[!t]
  \centering
  \small
  \caption{The test win rate of different MARL methods with our CADP framework and the CTDE framework.}
    \begin{tabular}{cccc}
    \toprule
    Method & 5m\_vs\_6m & corridor & 3s5z\_vs\_3s6z \\
    \midrule
    VDN (CTDE) & 0.54 $\pm$ 0.09  & 0.65 $\pm$ 0.32 & 0.25 $\pm$ 0.18 \\
    VDN (CADP) & \textbf{0.66 $\pm$ 0.07} & \textbf{0.72 $\pm$ 0.51}  &  \textbf{0.85 $\pm$ 0.20} \\ \midrule
    QMIX (CTDE) &  0.43 $\pm$ 0.13 & 0.70 $\pm$ 0.35 & 0.24 $\pm$ 0.36 \\
    QMIX (CADP) &  \textbf{0.68 $\pm$ 0.08} & \textbf{0.84 $\pm$ 0.03} & \textbf{0.93 $\pm$ 0.03} \\ \midrule
    QPLEX (CTDE) & 0.57 $\pm$ 0.13 & 0.20 $\pm$ 0.12 &  0.08 $\pm$ 0.11 \\
    QPLEX (CADP) & \textbf{0.73 $\pm$ 0.04} &  \textbf{0.37 $\pm$ 0.36} &  \textbf{0.96 $\pm$ 0.02} \\ \midrule
    MAPPO (CTDE) & 0.85$\pm$ 0.07 & 0.96 $\pm$ 0.03 & 0.35 $\pm$ 0.39\ \\
    MAPPO (CADP) & \textbf{0.97$\pm$ 0.03} & \textbf{0.98 $\pm$ 0.02} & \textbf{0.90 $\pm$ 0.16} \\
    \bottomrule
    \end{tabular}%
  \label{tab:example}%
   \vspace{-0.2cm}
\end{table}%

  \begin{figure}[!t]
    \centering
    \subfloat{\includegraphics[width=0.23\textwidth]{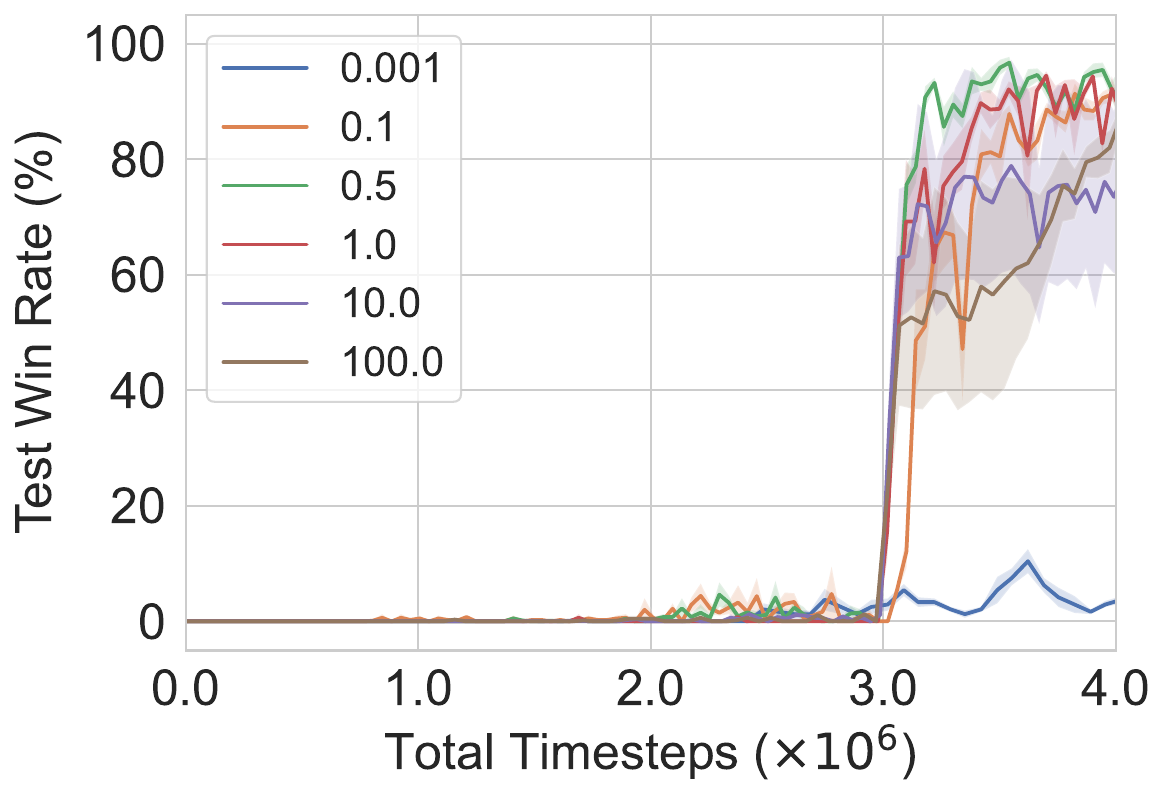}}
    \subfloat{\includegraphics[width=0.23\textwidth]{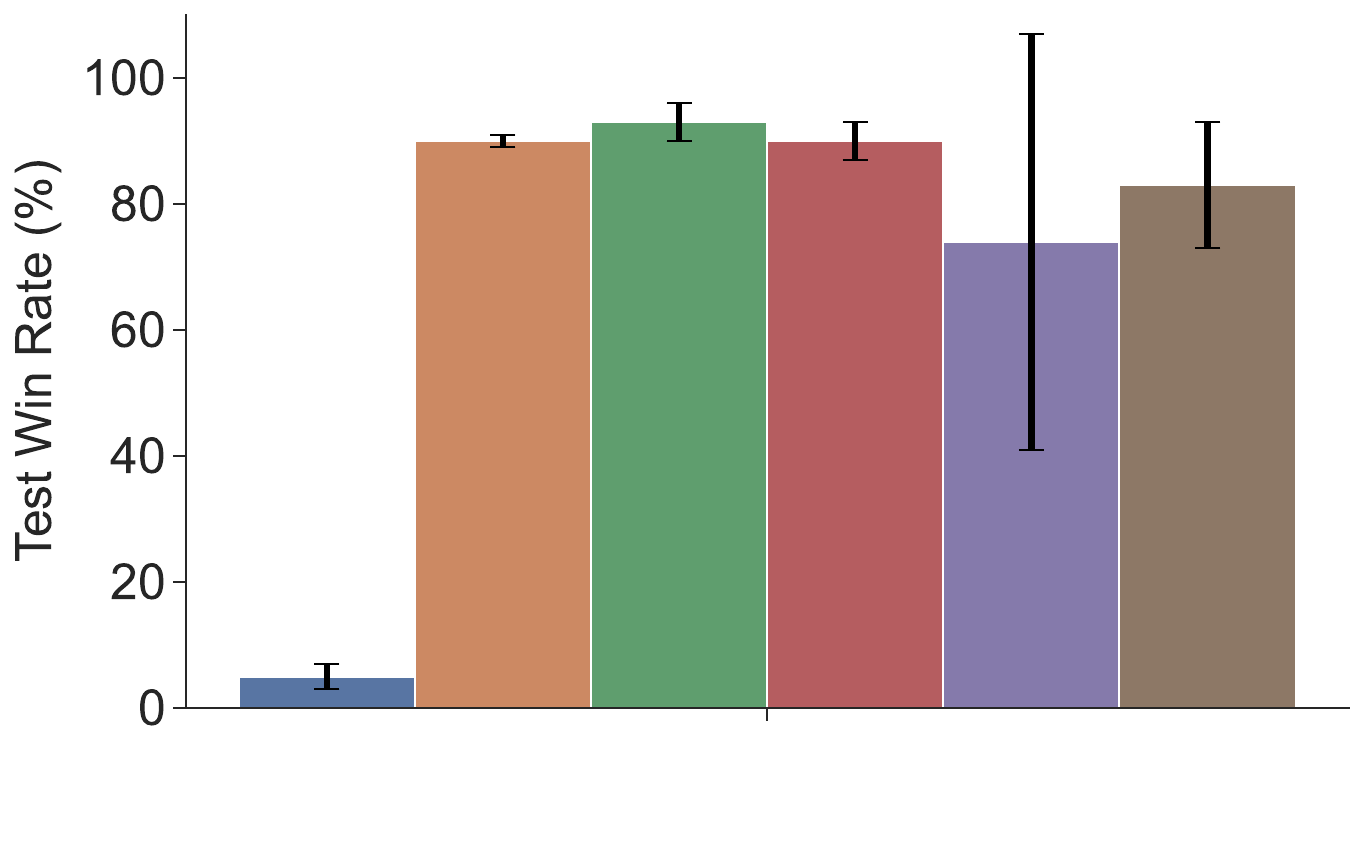}}
    \caption{Ablation study on different coefficients $\alpha$. The left part is the learning curves for 4M timesteps and the right part is the average test win rate of last 0.1M timesteps.}
    \label{fig:alpha}
     \vspace{-0.3cm}
  \end{figure}

\begin{figure}[!t]
    \centering
    \subfloat{\includegraphics[width=0.23\textwidth]{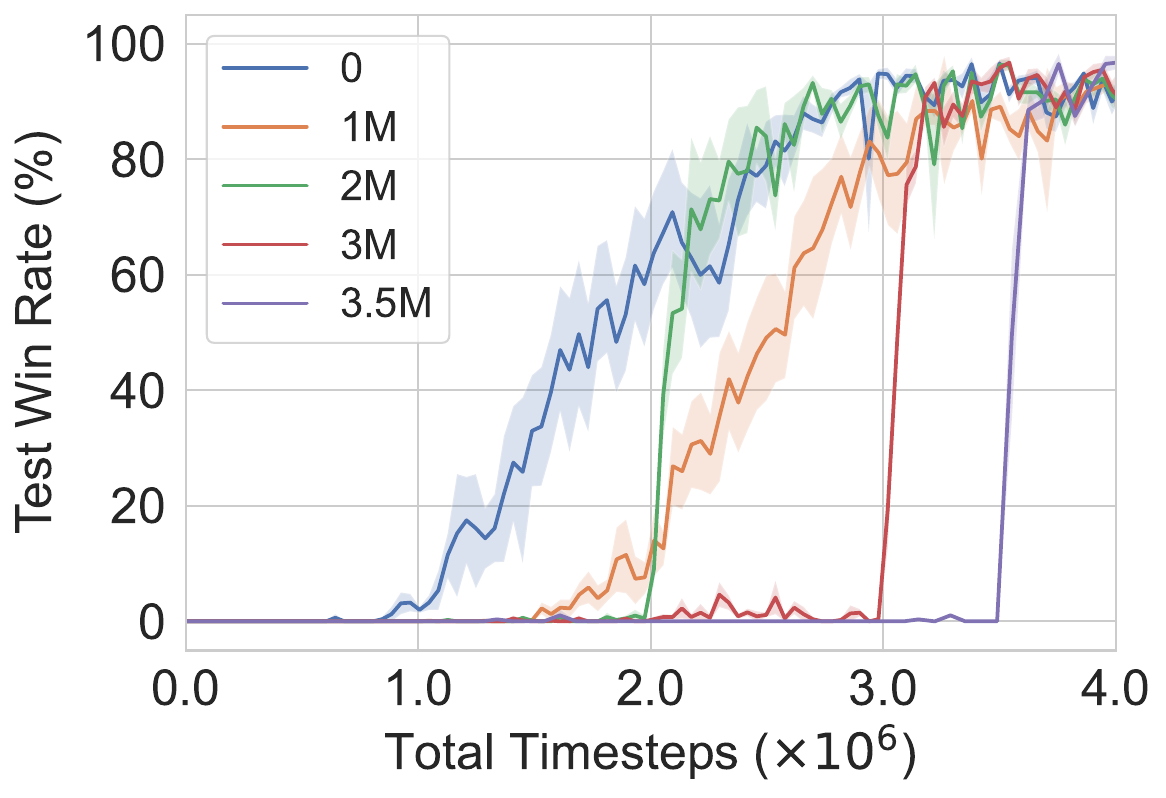}}
    \subfloat{\includegraphics[width=0.23\textwidth]{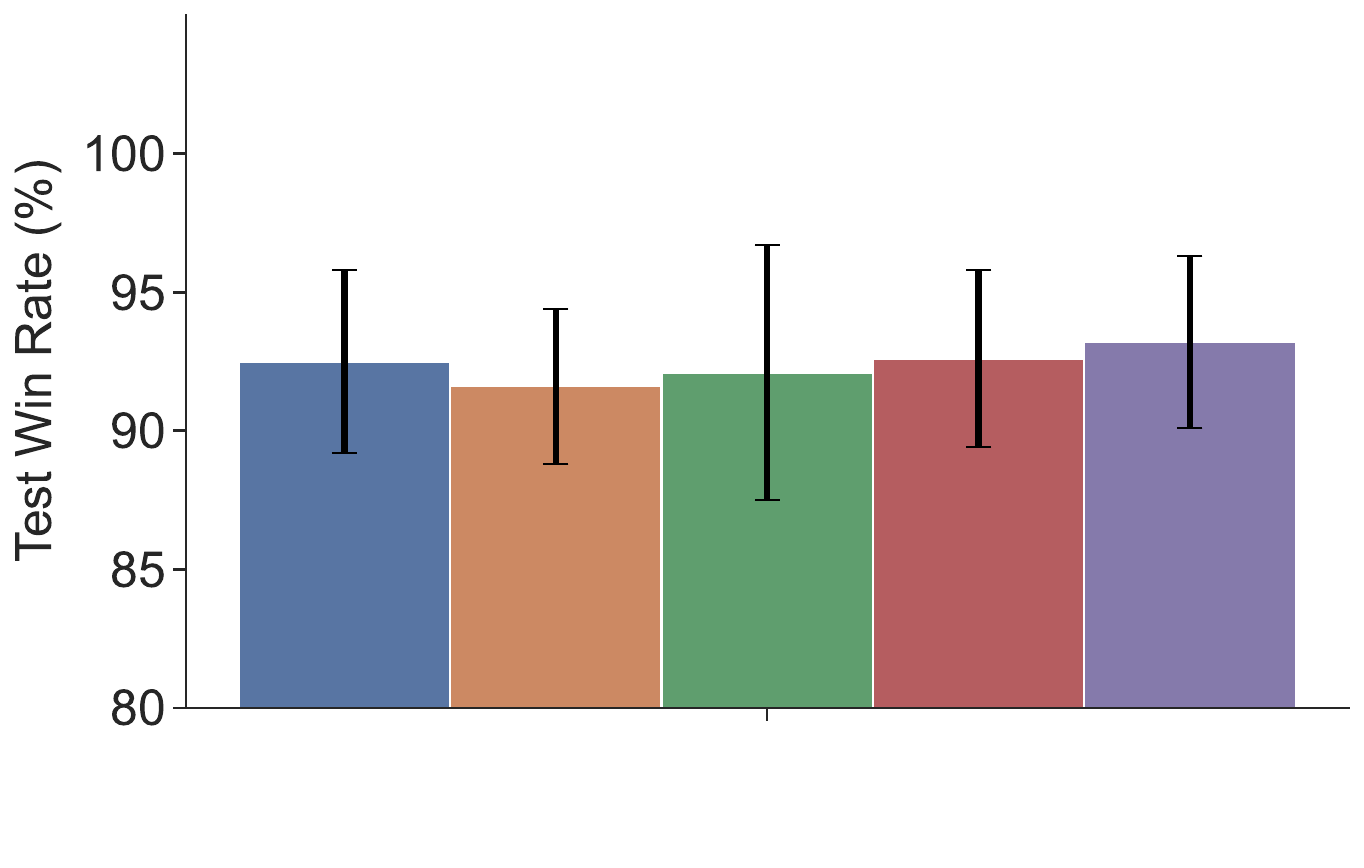}}
    \caption{Ablation study on different timestep $T$. The left part is the learning curves for 4M timesteps and the right part is the average test win rate of last 0.1M timesteps.   }
    \label{fig:T}
      \vspace{-0.3cm}
  \end{figure}

\textbf{Different hyperparameter.} We examine the effect of  the coefficient $\alpha$  in \emph{3s5z\_vs\_3s6z} scenarios in Figure~\ref{fig:alpha}. We observe
that there is not much difference in performance except the coefficient is too small~($\alpha=0.001$) to affect the training. Even when we set $\alpha = 100.0$, The final performance is also not far behind the best~($\alpha=0.5$). This means our method is very robust for the coefficient $\alpha$.
In addition, we consider five $T$ values: 0, 1M, 2M , 3M, and 3.5M. Figure~\ref{fig:T} demonstrates that using smaller $T$ for training, the performance of decentralized starts to improve earlier but  taked longer time from the beginning of improvement to convergence which may waste some computing resources while using bigger $T$ for training, the performance of decentralized model  improves more quickly but the convergence time may delay. In general, setting $T=3M$ is a balanced choice for both sides. 
The experiment also shows that different values have little impact on the final performance, which may be because $\mathcal{L}_{TD}$ is always dominant in training before the convergence of the centralized model.



    

\section{Conclusion}\label{conclusion}
In this paper, we argue the traditional CTDE framework is  not centralized enough, since  it falls short of fully utilizing global information for training, leading to an inefficient exploration of the joint-policy, and resulting in performance degration. Thus, we propose a novel Centralized Advising and Decentralized Pruning framework, termed as CADP, to enhance basic CTDE with global cooperative information. 
It is noting that our focus is not to design a new communication method. Our main contribution is adopting agent communication to enhance the basic CTDE framework for fully centralized training and still guarantees the independent policy for decentralized execution.   As CADP provides a light yet efficient training framework for many traditional CTDE methods,
we believe there will be more discussion and exploration under CADP framework. 

\section{Acknowledgments}
This work was supported in part by the CCF-Baidu Open Fund under Grant No. CCF-BAIDU OF202410, in part by the Hangzhou Joint Funds of the Zhejiang Provincial Natural Science Foundation of China under Grant No. LHZSD24F020001, in part by the Zhejiang Province High-Level Talents Special Support Program ``Leading Talent of Technological Innovation of Ten-Thousands Talents Program'' under Grant No. 2022R52046, in part by the Zhejiang Provincial Natural Science Foundation of China under Grant No. LMS25F020012, and in part by the advanced computing resources provided by the Supercomputing Center of Hangzhou City University.
\appendix

\bibliographystyle{named}
\bibliography{ijcai25}

\begin{thebibliography}{}

\bibitem[\protect\citeauthoryear{Chen \bgroup \em et al.\egroup }{2024}]{PTDE}
Yiqun Chen, Hangyu Mao, Tianle Zhang, Shiguang Wu, Bin Zhang, Jianye Hao, Dong Li, Bin Wang, and Hongxing Chang.
\newblock {PTDE:} personalized training with distillated execution for multi-agent reinforcement learning.
\newblock In {\em International Joint Conference on Artificial Intelligence}, 2024.

\bibitem[\protect\citeauthoryear{Claus and Boutilier}{1998}]{centralized}
Caroline Claus and Craig Boutilier.
\newblock The dynamics of reinforcement learning in cooperative multiagent systems.
\newblock In {\em AAAI Conference on Artificial Intelligence}, 1998.

\bibitem[\protect\citeauthoryear{Das \bgroup \em et al.\egroup }{2019}]{das2019tarmac}
Abhishek Das, Th{\'e}ophile Gervet, Joshua Romoff, Dhruv Batra, Devi Parikh, Mike Rabbat, and Joelle Pineau.
\newblock Tarmac: Targeted multi-agent communication.
\newblock In {\em International Conference on Machine Learning}, 2019.

\bibitem[\protect\citeauthoryear{Ding \bgroup \em et al.\egroup }{2020}]{I2C}
Ziluo Ding, Tiejun Huang, and Zongqing Lu.
\newblock Learning individually inferred communication for multi-agent cooperation.
\newblock In {\em Annual Conference on Neural Information Processing Systems}, 2020.

\bibitem[\protect\citeauthoryear{Ellis \bgroup \em et al.\egroup }{2023}]{ellis2022smacv2}
Benjamin Ellis, Skander Moalla, Mikayel Samvelyan, Mingfei Sun, Anuj Mahajan, Jakob~N Foerster, and Shimon Whiteson.
\newblock Smacv2: An improved benchmark for cooperative multi-agent reinforcement learning.
\newblock {\em Annual Conference on Neural Information Processing Systems}, 2023.

\bibitem[\protect\citeauthoryear{Foerster \bgroup \em et al.\egroup }{2018}]{COMA}
Jakob~N. Foerster, Gregory Farquhar, Triantafyllos Afouras, Nantas Nardelli, and Shimon Whiteson.
\newblock Counterfactual multi-agent policy gradients.
\newblock In {\em AAAI Conference on Artificial Intelligence}, 2018.

\bibitem[\protect\citeauthoryear{Fu \bgroup \em et al.\egroup }{2022}]{fu2022revisiting}
Wei Fu, Chao Yu, Zelai Xu, Jiaqi Yang, and Yi~Wu.
\newblock Revisiting some common practices in cooperative multi-agent reinforcement learning.
\newblock In {\em International Conference on Machine Learning}, 2022.

\bibitem[\protect\citeauthoryear{Hong \bgroup \em et al.\egroup }{2022}]{dagger}
Yitian Hong, Yaochu Jin, and Yang Tang.
\newblock Rethinking individual global max in cooperative multi-agent reinforcement learning.
\newblock In {\em Annual Conference on Neural Information Processing Systems}, 2022.

\bibitem[\protect\citeauthoryear{Hu \bgroup \em et al.\egroup }{2024}]{iclr2024graph}
Shengchao Hu, Li~Shen, Ya~Zhang, and Dacheng Tao.
\newblock Learning multi-agent communication from graph modeling perspective.
\newblock In {\em International Conference on Learning Representations}, 2024.

\bibitem[\protect\citeauthoryear{Iqbal and Sha}{2019}]{MAAC}
Shariq Iqbal and Fei Sha.
\newblock Actor-attention-critic for multi-agent reinforcement learning.
\newblock In {\em International Conference on Machine Learning}, 2019.

\bibitem[\protect\citeauthoryear{Jiang and Lu}{2018}]{ATOC}
Jiechuan Jiang and Zongqing Lu.
\newblock Learning attentional communication for multi-agent cooperation.
\newblock In {\em Annual Conference on Neural Information Processing Systems}, 2018.

\bibitem[\protect\citeauthoryear{Jiang \bgroup \em et al.\egroup }{2021}]{jiang2021action}
Haobo Jiang, Jin Xie, and Jian Yang.
\newblock Action candidate based clipped double q-learning for discrete and continuous action tasks.
\newblock In {\em Proceedings of the AAAI conference on artificial intelligence}, 2021.

\bibitem[\protect\citeauthoryear{Jiang \bgroup \em et al.\egroup }{2024}]{iclr2024non-stationary}
Haozhe Jiang, Qiwen Cui, Zhihan Xiong, Maryam Fazel, and Simon~S. Du.
\newblock A black-box approach for non-stationary multi-agent reinforcement learning.
\newblock In {\em International Conference on Learning Representations}, 2024.

\bibitem[\protect\citeauthoryear{Kapoor \bgroup \em et al.\egroup }{2024}]{kapoor2024}
Aditya Kapoor, Benjamin Freed, Howie Choset, and Jeff Schneider.
\newblock Assigning credit with partial reward decoupling in multi-agent proximal policy optimization, 2024.

\bibitem[\protect\citeauthoryear{Kuba \bgroup \em et al.\egroup }{2022}]{HATRPO}
Jakub~Grudzien Kuba, Ruiqing Chen, Muning Wen, Ying Wen, Fanglei Sun, Jun Wang, and Yaodong Yang.
\newblock Trust region policy optimisation in multi-agent reinforcement learning.
\newblock In {\em International Conference on Learning Representations}, 2022.

\bibitem[\protect\citeauthoryear{Kurach \bgroup \em et al.\egroup }{2020}]{GRF}
Karol Kurach, Anton Raichuk, Piotr Stanczyk, Michal Zajac, Olivier Bachem, Lasse Espeholt, Carlos Riquelme, Damien Vincent, Marcin Michalski, Olivier Bousquet, and Sylvain Gelly.
\newblock Google research football: A novel reinforcement learning environment.
\newblock In {\em AAAI Conference on Artificial Intelligence}, 2020.

\bibitem[\protect\citeauthoryear{Li \bgroup \em et al.\egroup }{2021}]{sharpley_credit}
Jiahui Li, Kun Kuang, Baoxiang Wang, Furui Liu, Long Chen, Fei Wu, and Jun Xiao.
\newblock Shapley counterfactual credits for multi-agent reinforcement learning.
\newblock In {\em ACM SIGKDD Conference on Knowledge Discovery and Data Mining}, 2021.

\bibitem[\protect\citeauthoryear{Li \bgroup \em et al.\egroup }{2023}]{li2022ace}
Chuming Li, Jie Liu, Yinmin Zhang, Yuhong Wei, Yazhe Niu, Yaodong Yang, Yu~Liu, and Wanli Ouyang.
\newblock Ace: Cooperative multi-agent q-learning with bidirectional action-dependency.
\newblock In {\em AAAI Conference on Artificial Intelligence}, 2023.

\bibitem[\protect\citeauthoryear{Liu \bgroup \em et al.\egroup }{2023}]{CIA}
Shunyu Liu, Yihe Zhou, Jie Song, Tongya Zheng, Kaixuan Chen, Tongtian Zhu, Zunlei Feng, and Mingli Song.
\newblock Contrastive identity-aware learning for multi-agent value decomposition.
\newblock In {\em AAAI Conference on Artificial Intelligence}, pages 11595--11603, 2023.

\bibitem[\protect\citeauthoryear{Liu \bgroup \em et al.\egroup }{2024}]{liu2024OPT}
Shunyu Liu, Jie Song, Yihe Zhou, Na~Yu, Kaixuan Chen, Zunlei Feng, and Mingli Song.
\newblock Interaction pattern disentangling for multi-agent reinforcement learning.
\newblock {\em IEEE Transactions on Pattern Analysis and Machine Intelligence}, 2024.

\bibitem[\protect\citeauthoryear{Lo \bgroup \em et al.\egroup }{2024}]{iclr2024contrastive}
Yat~Long Lo, Biswa Sengupta, Jakob Foerster, and Michael Noukhovitch.
\newblock Learning to communicate using contrastive learning.
\newblock In {\em International Conference on Learning Representations}, 2024.

\bibitem[\protect\citeauthoryear{Lowe \bgroup \em et al.\egroup }{2017}]{MADDPG}
Ryan Lowe, Yi~Wu, Aviv Tamar, Jean Harb, Pieter Abbeel, and Igor Mordatch.
\newblock Multi-agent actor-critic for mixed cooperative-competitive environments.
\newblock In {\em Annual Conference on Neural Information Processing Systems}, 2017.

\bibitem[\protect\citeauthoryear{Mao \bgroup \em et al.\egroup }{2020a}]{GACML}
Hangyu Mao, Zhengchao Zhang, Zhen Xiao, Zhibo Gong, and Yan Ni.
\newblock Learning agent communication under limited bandwidth by message pruning.
\newblock In {\em AAAI Conference on Artificial Intelligence}, 2020.

\bibitem[\protect\citeauthoryear{Mao \bgroup \em et al.\egroup }{2020b}]{DAACMP}
Hangyu Mao, Zhengchao Zhang, Zhen Xiao, Zhibo Gong, and Yan Ni.
\newblock Learning multi-agent communication with double attentional deep reinforcement learning.
\newblock {\em International Joint Conference on Autonomous Agents and Multi-agent Systems}, 2020.

\bibitem[\protect\citeauthoryear{Mnih \bgroup \em et al.\egroup }{2015}]{DQN15}
Volodymyr Mnih, Koray Kavukcuoglu, David Silver, Andrei~A. Rusu, et~al.
\newblock Human-level control through deep reinforcement learning.
\newblock {\em Nature}, 518(7540):529--533, 2015.

\bibitem[\protect\citeauthoryear{Nayak \bgroup \em et al.\egroup }{2023}]{nayak2023scalable}
Siddharth Nayak, Kenneth Choi, Wenqi Ding, Sydney Dolan, Karthik Gopalakrishnan, and Hamsa Balakrishnan.
\newblock Scalable multi-agent reinforcement learning through intelligent information aggregation, 2023.

\bibitem[\protect\citeauthoryear{Phan \bgroup \em et al.\egroup }{2023}]{messy_smac}
Thomy Phan, Fabian Ritz, Philipp Altmann, Maximilian Zorn, Jonas Nüßlein, Michael Kölle, Thomas Gabor, and Claudia Linnhoff-Popien.
\newblock Attention-based recurrence for multi-agent reinforcement learning under stochastic partial observability.
\newblock In {\em International Conference on Machine Learning}, 2023.

\bibitem[\protect\citeauthoryear{Qing \bgroup \em et al.\egroup }{2024}]{A2PO_yunpeng}
Yunpeng Qing, Shunyu Liu, Jingyuan Cong, Kaixuan Chen, Yihe Zhou, and Mingli Song.
\newblock {A2PO:} towards effective offline reinforcement learning from an advantage-aware perspective.
\newblock In {\em Annual Conference on Neural Information Processing Systems}, 2024.

\bibitem[\protect\citeauthoryear{Rashid \bgroup \em et al.\egroup }{2018}]{QMIX}
Tabish Rashid, Mikayel Samvelyan, Christian~Schr{\"{o}}der de~Witt, Gregory Farquhar, Jakob~N. Foerster, and Shimon Whiteson.
\newblock {QMIX:} monotonic value function factorisation for deep multi-agent reinforcement learning.
\newblock In {\em International Conference on Machine Learning}, 2018.

\bibitem[\protect\citeauthoryear{Rashid \bgroup \em et al.\egroup }{2020}]{WQMIX}
Tabish Rashid, Gregory Farquhar, Bei Peng, and Shimon Whiteson.
\newblock Weighted {QMIX:} expanding monotonic value function factorisation for deep multi-agent reinforcement learning.
\newblock In {\em Annual Conference on Neural Information Processing Systems}, 2020.

\bibitem[\protect\citeauthoryear{Samvelyan \bgroup \em et al.\egroup }{2019}]{SMAC}
Mikayel Samvelyan, Tabish Rashid, Christian~Schr{\"{o}}der de~Witt, Gregory Farquhar, Nantas Nardelli, Tim G.~J. Rudner, Chia{-}Man Hung, Philip H.~S. Torr, Jakob~N. Foerster, and Shimon Whiteson.
\newblock The starcraft multi-agent challenge.
\newblock In {\em International Joint Conference on Autonomous Agents and Multi-agent Systems}, 2019.

\bibitem[\protect\citeauthoryear{Schulman \bgroup \em et al.\egroup }{2017}]{2017PPO}
John Schulman, Filip Wolski, Prafulla Dhariwal, Alec Radford, and Oleg Klimov.
\newblock Proximal policy optimization algorithms.
\newblock {\em arXiv preprint arXiv:1707.06347}, 2017.

\bibitem[\protect\citeauthoryear{Son \bgroup \em et al.\egroup }{2019}]{QTRAN}
Kyunghwan Son, Daewoo Kim, Wan~Ju Kang, David Hostallero, and Yung Yi.
\newblock {QTRAN:} learning to factorize with transformation for cooperative multi-agent reinforcement learning.
\newblock In {\em International Conference on Machine Learning}, 2019.

\bibitem[\protect\citeauthoryear{Stenning \bgroup \em et al.\egroup }{2006}]{stenning2006introduction}
Keith Stenning, Jo~Calder, and Alex Lascarides.
\newblock {\em Introduction to cognition and communication}.
\newblock MIT Press, 2006.

\bibitem[\protect\citeauthoryear{Su \bgroup \em et al.\egroup }{2022}]{su2022ma2ql}
Kefan Su, Siyuan Zhou, Chuang Gan, Xiangjun Wang, and Zongqing Lu.
\newblock Ma2ql: A minimalist approach to fully decentralized multi-agent reinforcement learning.
\newblock {\em arXiv preprint arXiv:2209.08244}, 2022.

\bibitem[\protect\citeauthoryear{Sukhbaatar \bgroup \em et al.\egroup }{2016}]{CommNet}
Sainbayar Sukhbaatar, Rob Fergus, et~al.
\newblock Learning multiagent communication with backpropagation.
\newblock In {\em Annual Conference on Neural Information Processing Systems}, 2016.

\bibitem[\protect\citeauthoryear{Sunehag \bgroup \em et al.\egroup }{2018}]{VDN}
Peter Sunehag, Guy Lever, Audrunas Gruslys, Wojciech~Marian Czarnecki, et~al.
\newblock Value-decomposition networks for cooperative multi-agent learning based on team reward.
\newblock In {\em International Joint Conference on Autonomous Agents and Multi-agent Systems}, 2018.

\bibitem[\protect\citeauthoryear{Sutton and Barto}{2018}]{sutton2018reinforcement}
Richard~S Sutton and Andrew~G Barto.
\newblock {\em Reinforcement learning: An introduction}.
\newblock MIT press, 2018.

\bibitem[\protect\citeauthoryear{Tan}{1993}]{iql}
Ming Tan.
\newblock Multi-agent reinforcement learning: Independent vs. cooperative agents.
\newblock In {\em International Conference on Machine Learning}, 1993.

\bibitem[\protect\citeauthoryear{Vaswani \bgroup \em et al.\egroup }{2017}]{attention}
Ashish Vaswani, Noam Shazeer, Niki Parmar, Jakob Uszkoreit, Llion Jones, Aidan~N Gomez, \L~ukasz Kaiser, and Illia Polosukhin.
\newblock Attention is all you need.
\newblock In {\em Annual Conference on Neural Information Processing Systems}, 2017.

\bibitem[\protect\citeauthoryear{Vinyals \bgroup \em et al.\egroup }{2019}]{AlphaStar}
Oriol Vinyals, Igor Babuschkin, Wojciech~M Czarnecki, Micha{\"e}l Mathieu, et~al.
\newblock Grandmaster level in starcraft ii using multi-agent reinforcement learning.
\newblock {\em Nature}, 575(7782):350--354, 2019.

\bibitem[\protect\citeauthoryear{Wang \bgroup \em et al.\egroup }{2020a}]{Bottleneck}
Rundong Wang, Xu~He, Runsheng Yu, Wei Qiu, Bo~An, and Zinovi Rabinovich.
\newblock Learning efficient multi-agent communication: An information bottleneck approach.
\newblock In {\em International Conference on Machine Learning}, 2020.

\bibitem[\protect\citeauthoryear{Wang \bgroup \em et al.\egroup }{2020b}]{IMAC}
Rundong Wang, Xu~He, Runsheng Yu, Wei Qiu, Bo~An, and Zinovi Rabinovich.
\newblock Learning efficient multi-agent communication: An information bottleneck approach.
\newblock In {\em International Conference on Machine Learning}, 2020.

\bibitem[\protect\citeauthoryear{Wang \bgroup \em et al.\egroup }{2020c}]{NDQ}
Tonghan Wang, Jianhao Wang, Chongyi Zheng, and Chongjie Zhang.
\newblock Learning nearly decomposable value functions via communication minimization.
\newblock In {\em International Conference on Learning Representations}, 2020.

\bibitem[\protect\citeauthoryear{Wang \bgroup \em et al.\egroup }{2021}]{QPLEX}
Jianhao Wang, Zhizhou Ren, Terry Liu, Yang Yu, and Chongjie Zhang.
\newblock {QPLEX:} duplex dueling multi-agent q-learning.
\newblock In {\em International Conference on Learning Representations}, 2021.

\bibitem[\protect\citeauthoryear{Wang \bgroup \em et al.\egroup }{2023}]{MACPF}
Jiangxing Wang, Deheng Ye, and Zongqing Lu.
\newblock More centralized training, still decentralized execution: Multi-agent conditional policy factorization.
\newblock In {\em International Conference on Learning Representations}, 2023.

\bibitem[\protect\citeauthoryear{Wu \bgroup \em et al.\egroup }{2020}]{wu2020multi}
Tong Wu, Pan Zhou, Kai Liu, Yali Yuan, Xiumin Wang, Huawei Huang, and Dapeng~Oliver Wu.
\newblock Multi-agent deep reinforcement learning for urban traffic light control in vehicular networks.
\newblock {\em IEEE Transactions on Vehicular Technology}, 69(8):8243--8256, 2020.

\bibitem[\protect\citeauthoryear{Xu \bgroup \em et al.\egroup }{2024}]{power_vipa}
Feiyang Xu, Shunyu Liu, Yunpeng Qing, Yihe Zhou, Yuwen Wang, and Mingli Song.
\newblock Temporal prototype-aware learning for active voltage control on power distribution networks.
\newblock In {\em ACM SIGKDD Conference on Knowledge Discovery and Data Mining}, 2024.

\bibitem[\protect\citeauthoryear{Yu \bgroup \em et al.\egroup }{2022}]{MAPPO}
Chao Yu, Akash Velu, Eugene Vinitsky, Yu~Wang, Alexandre Bayen, and Yi~Wu.
\newblock The surprising effectiveness of ppo in cooperative, multi-agent games.
\newblock In {\em Annual Conference on Neural Information Processing Systems}, 2022.

\bibitem[\protect\citeauthoryear{Yuan \bgroup \em et al.\egroup }{2022}]{MAIC}
Lei Yuan, Jianhao Wang, Fuxiang Zhang, Chenghe Wang, Zongzhang Zhang, Yang Yu, and Chongjie Zhang.
\newblock Multi-agent incentive communication via decentralized teammate modeling.
\newblock In {\em AAAI Conference on Artificial Intelligence}, 2022.

\bibitem[\protect\citeauthoryear{Zhao \bgroup \em et al.\egroup }{2022}]{CTDS}
Jian Zhao, Xunhan Hu, Mingyu Yang, Wengang Zhou, Jiangcheng Zhu, and Houqiang Li.
\newblock {CTDS:} centralized teacher with decentralized student for multi-agent reinforcement learning.
\newblock {\em IEEE Transactions on Games}, 16(1):140--150, 2022.

\end{thebibliography}
\clearpage

\twocolumn
\appendix                 
\renewcommand\thefigure{\thesection-\arabic{figure}}
\renewcommand\thetable {\thesection-\arabic{table}}
\renewcommand\theequation{\thesection-\arabic{equation}}
\setcounter{figure}{0}    
\setcounter{table}{0}
\setcounter{equation}{0}

\section{Additional Related Work}~\label{add_rela}

Early learning frameworks in cooperative MARL can be mainly categorized into two classes: centralized learning and decentralized learning. Centralized learning~\cite{centralized} learns a single policy to directly produce the joint actions of all agents, but in many real-world scenarios, the partial observability and communication constraints among agents limit the feasibility of this framework. In decentralized learning~\cite{iql}, each agent can optimize its reward independently. Thus, it can be used in most of real-world scenarios. However, non-stationarity  still remains a significant challenge in decentralized learning~\cite{su2022ma2ql}. CTDE combines the advantages of these two frameworks, where agents can enjoy centralized training with additional global information and deliver actions only based on their local observation in a decentralized way.

Value Decomposition~(VD) is a well-established technique for cooperative MARL problems under CTDE framework. VD is supposed to decompose the joint value into the individual value properly. To achieve efficient VD, it is necessary to satisfy the Individual-Global-Max~(IGM) principle that the global optimal action should be consistent with the collection of individual optimal actions of agents~\cite{QTRAN}. Following the IGM principle, VDN~\cite{VDN} proposes to represent the joint value function as a sum of individual value functions, while QMIX~\cite{QMIX} changes the simple sum function of VDN to a deep neural network that satisfies monotonicity constraints. 
To further alleviate the risk of suboptimal results, WQMIX~\cite{WQMIX} improves QMIX by a weighted projection that allows more emphasis to be placed on underestimated actions. However, both VDN and QMIX use stricter constraints than IGM, which limit their representation capability for joint action-value function classes. Thus, QTRAN~\cite{QTRAN} and QPLEX~\cite{QPLEX} propose some novel methods whose constraints is looser than VDN and QMIX, but still satisfy the IGM~principle. 

Several conventional CTDE-based methods, such as COMA~\cite{COMA} and MADDPG~\cite{MADDPG},  adopt another algorithm called Policy Gradient~(PG), which learns local agent policies as the actor and learns a centralized value function as the mix module. Because of the on-policy fashion, PG methods are typically believed to be less sample efficient and therefore less applied in challenging tasks with limited computation resources~\cite{fu2022revisiting}. Whereas, MAPPO~\cite{MAPPO} and HAPPO~\cite{HATRPO} demonstrate that with appropriate implementation tricks of PPO algoritm~\cite{2017PPO}, PG method can achieve strong performance and sample efficiency in many cooperative MARL benchmarks compared to the VD method.
\begin{figure*}[!b]
    \centering
    \subfloat[5m\_vs\_6m]{\includegraphics[width=0.30\textwidth]{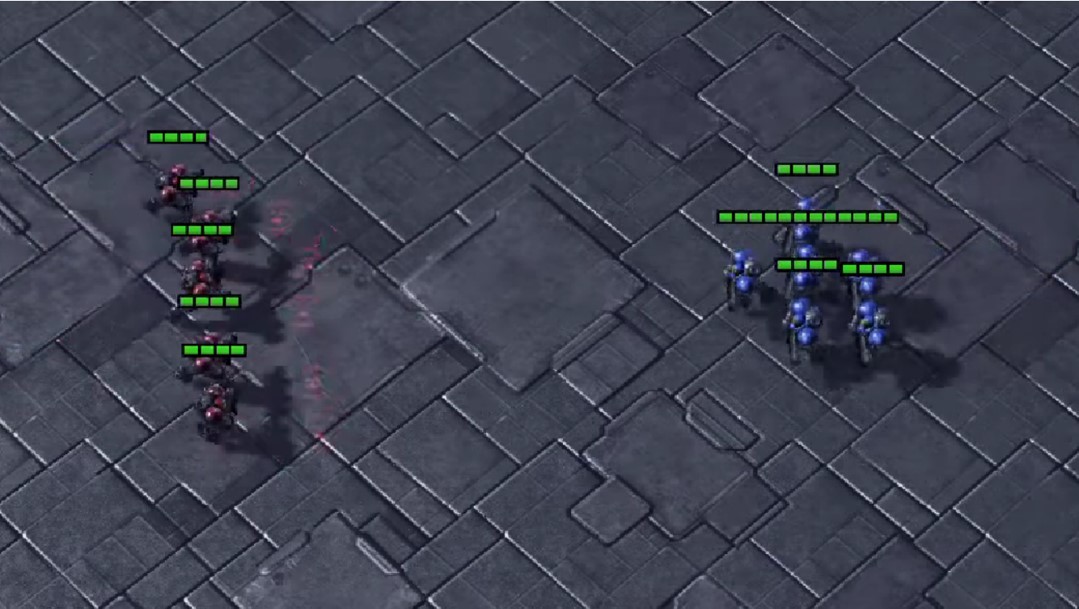}} \quad
    \subfloat[corridor]{\includegraphics[width=0.30\textwidth]{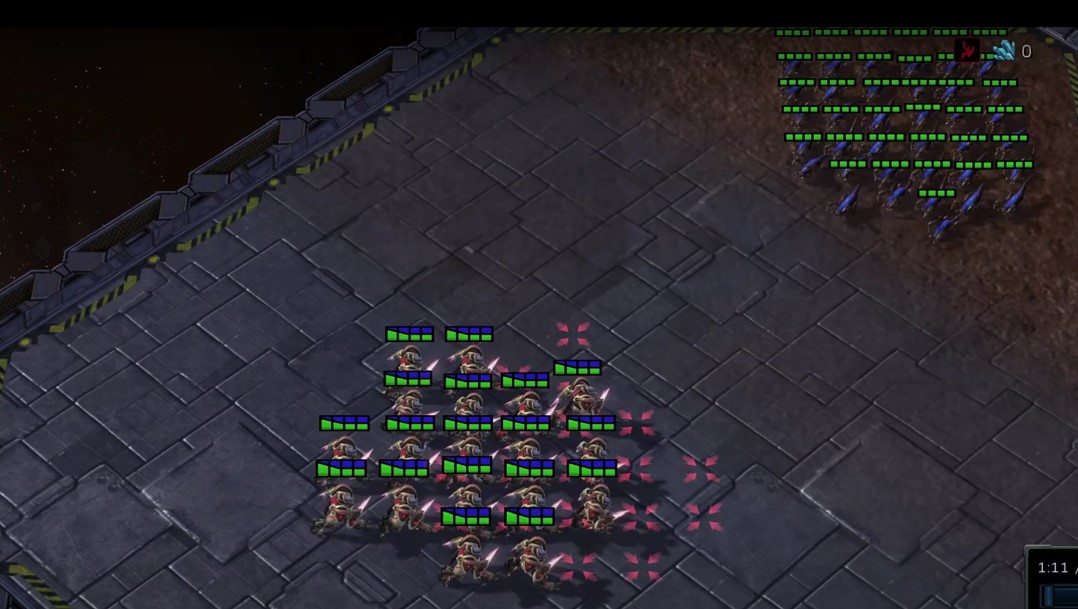}} \quad
    \subfloat[3s5z\_vs\_3s6z]{\includegraphics[width=0.30\textwidth]{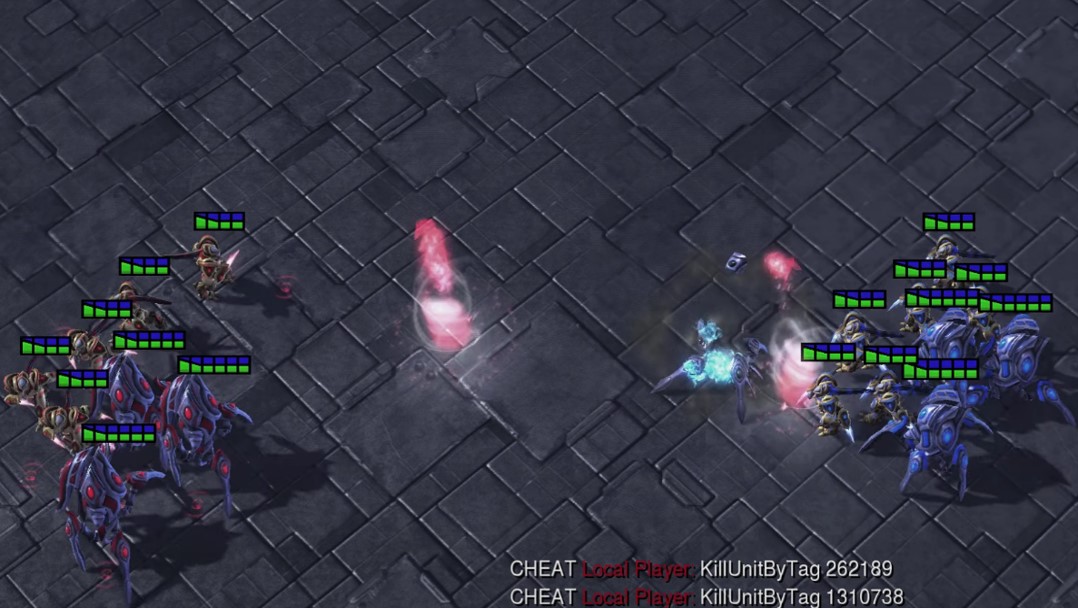}} \\
    \caption{The initial state of three SMAC scenarios.}
    \label{fig:sc2}
  \end{figure*}

\begin{figure*}[!b]
    \centering
    \subfloat[academy\_3\_vs\_1\_with\_keeper]{\includegraphics[width=0.33\textwidth]{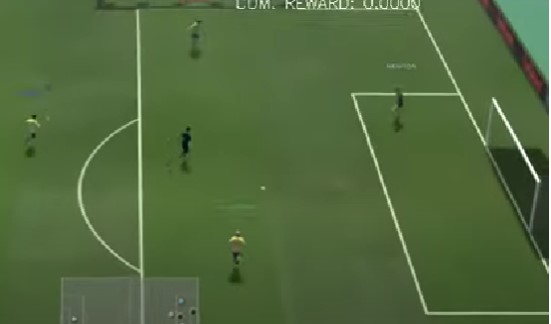}} \quad
    \subfloat[academy\_counterattack\_easy]{\includegraphics[width=0.33\textwidth]{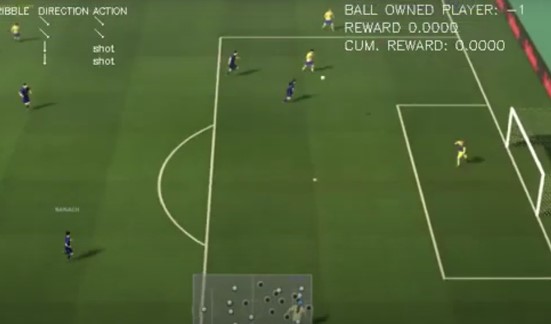}} \\
    \caption{Two scenarios of Google Research Football game.}
    \label{fig:grf_picture}
  \end{figure*}

\section{Detailed Hyperparameters} \label{hyperparameters}
 We adopt the common-used Python MARL framework (PyMARL)~\cite{SMAC} to implement our method and all baselines. The detailed hyperparameters in  the SMAC benchmark and GRF benchmark are given as follows, where the common hyperparameters across different methods are consistent for a fair comparison. We modify several hyperparameters in some difficult scenarios and also detail them below. All experimental results are illustrated with the mean and variance of the performance over five random seeds for a fair comparison. To make the results in figures clearer for readers, we adopt a 50\% confidence interval to plot the error bar. 
 
 Batches of 32 episodes are sampled from the replay buffer with the size of 5K for every training iteration. The target update interval is set to 200, and the discount factor is set to 0.99. We use the Adam Optimizer with a learning rate of $5 \times 10^{-4}$ for SMAC scenarios, while use the RMSprop Optimizer with a learning rate of $5 \times 10^{-4}$, a smoothing constant of 0.99, and with no momentum or weight decay for GRF scenarios. For exploration, $\epsilon$-greedy is used with $\epsilon$ annealed linearly from 1.0 to 0.05 over 50K training steps and kept constant for the rest of the training. In several super hard SMAC scenarios~(\emph{corridor}, \emph{3s5z\_vs\_3s6z}) that require more exploration. In this way, we extend the epsilon anneal time to 500K for all the compared methods. For the CADP module, the additional hyperparameters are the trade-off coefficient $\alpha$ of the pruning loss  $ \mathcal{L}_{p}$ and the timestep $T$ when  the pruning loss  $ \mathcal{L}_{p}$ is used. We search $\alpha$ over $\{0.001, 0.1, 0.5, 1.0, 10.0, 100.0\}$ in the SMAC benchmark. The coefficient $\alpha$ of the pruning loss  $ \mathcal{L}_{p}$  is set to 1.0, 0.5, and 0.5 for \emph{5m\_vs\_6m}, \emph{corridor} and \emph{3s5z\_vs\_3s6z}, respectively and the timestep $T$ is set to about three-quarters of the total timesteps. We search $\alpha$ over $\{ 0.5, 1.0, 10.0, 30.0, 50.0, 100.0\}$ in the GRF benchmark, while the coefficient $\alpha$ of the pruning loss  $ \mathcal{L}_{p}$ is set to 50.0 and 30.0 for \emph{3\_vs\_1\_with\_keeper} and \emph{counterattack\_easy}, respectively and the timestep $T$ is set to about three-fifths of the total timesteps. That is because GRF benchmark is harder than SMAC benchmark for model-pruning. In the ablation study, the coefficient $\alpha$ is consistent with the above.  
 
 Experiments are carried out on NVIDIA P5000 GPU.  Besides, in GRF scenarios, some methods are not supported by their original source code. So we follow the design in their papers and add some code to run GRF scenarios.

\section{Additional Experiments}\label{more_exp}

\begin{figure*}[!t]
    \centering
     \subfloat{\quad\;\includegraphics[width=0.6\textwidth]{fig/legend3_4.pdf}}\vspace{-0.8em}\\
     \addtocounter{subfigure}{-1}
    \subfloat[Sight Range of 9]{\includegraphics[width=0.33\textwidth]{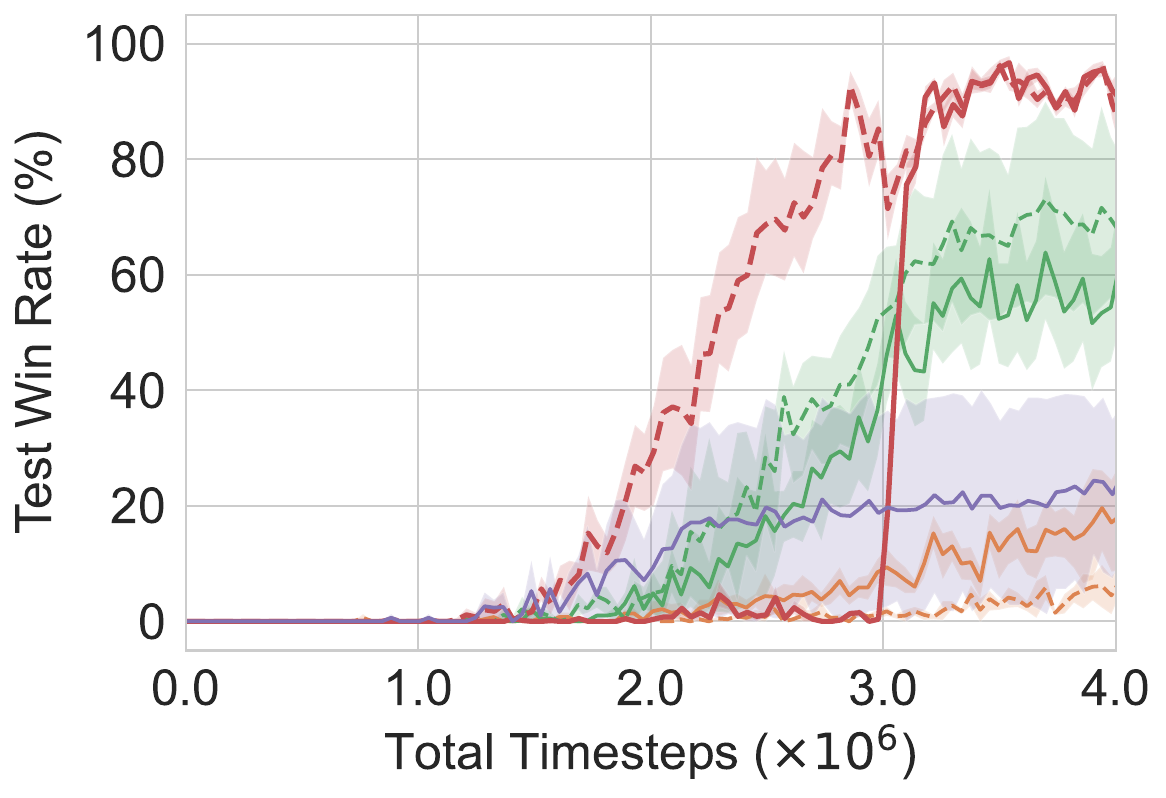}}
    \subfloat[Sight Range of 5]{\includegraphics[width=0.33\textwidth]{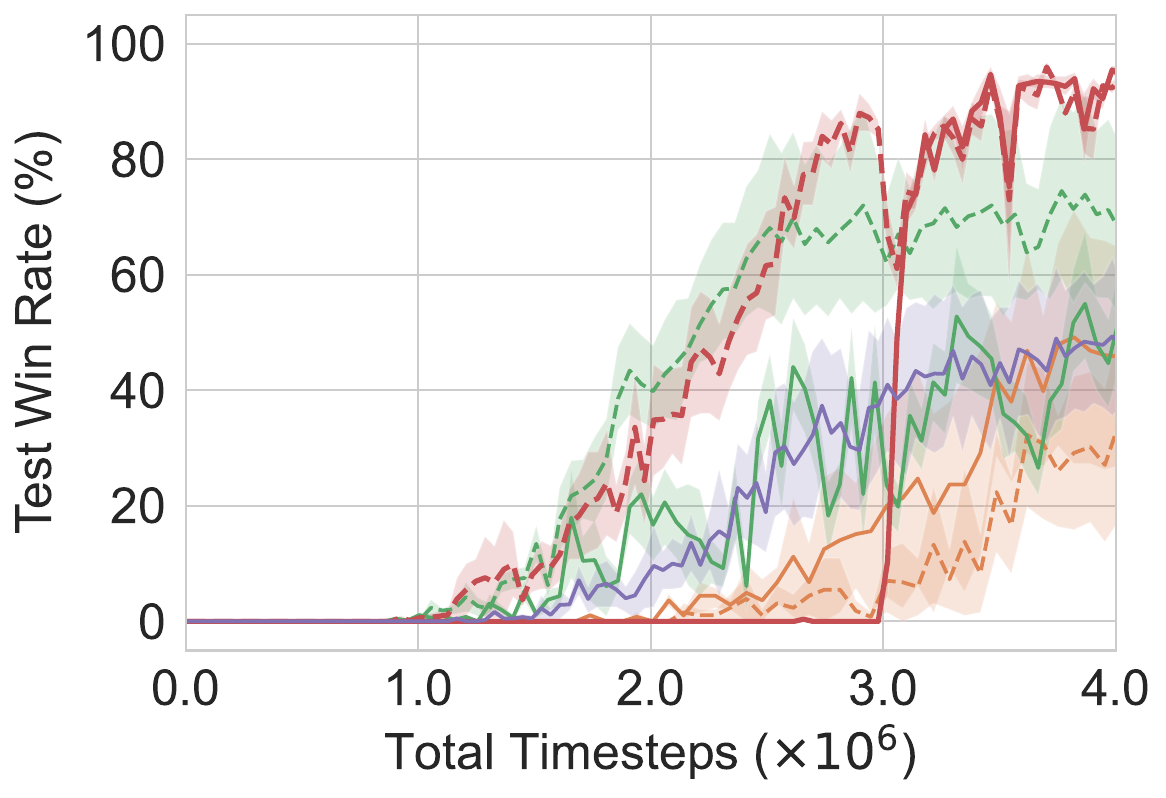}}
    \subfloat[Sight Range of 0]{\includegraphics[width=0.33\textwidth]{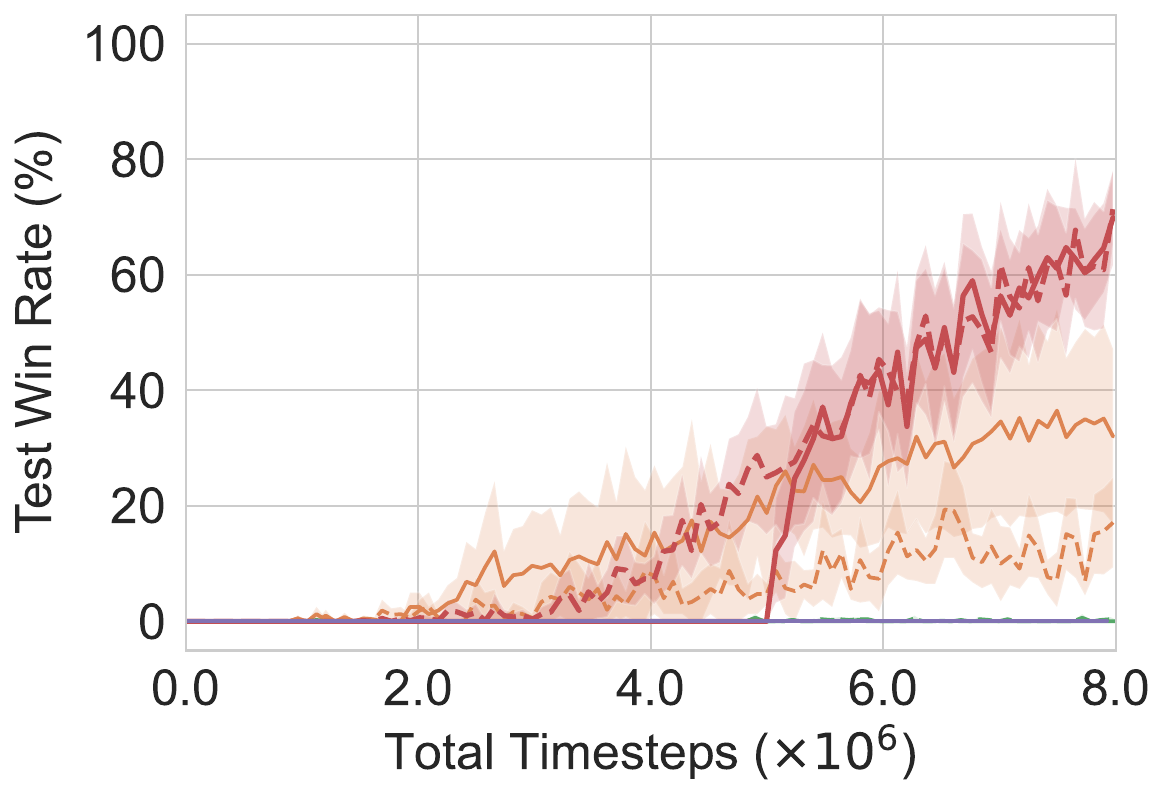}}
    \caption{Ablation study on different sight ranges.}
    \label{fig:sightrange}
  \end{figure*}

\begin{figure}[t]
  \centering
  
      \subfloat{\includegraphics[width=0.23\textwidth]{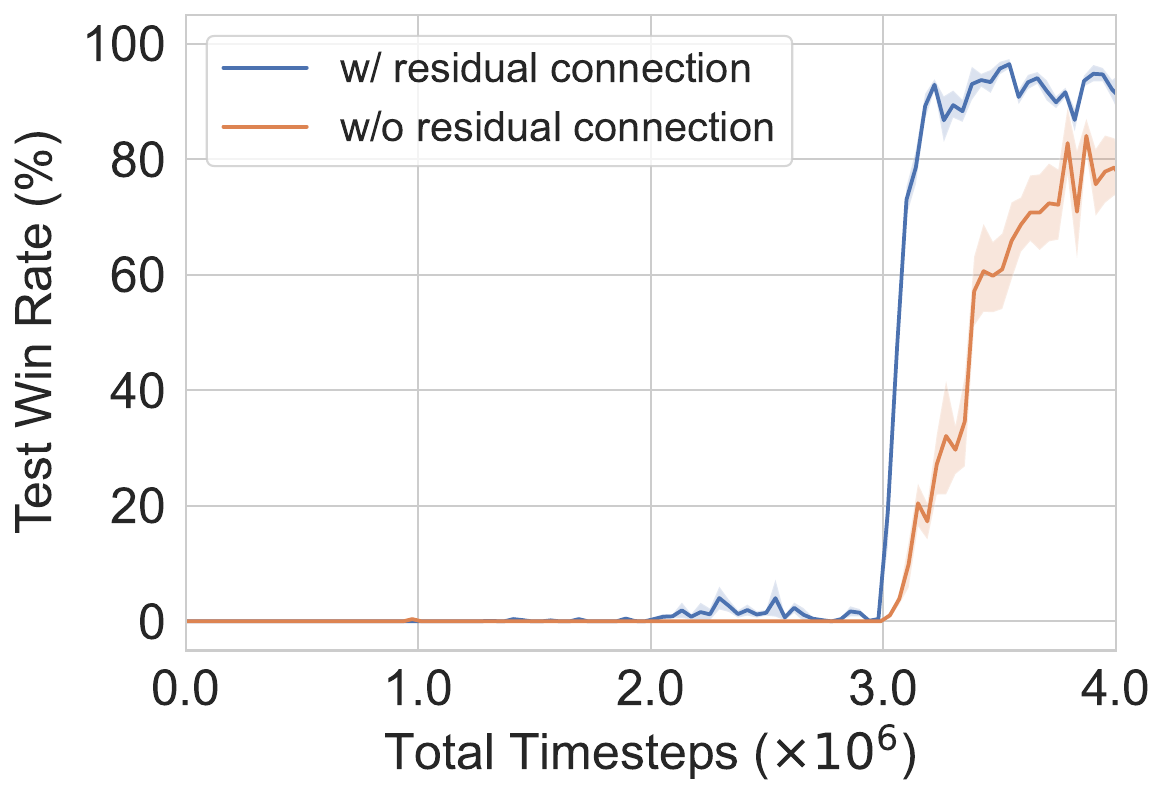}}
      \subfloat{\includegraphics[width=0.23\textwidth]{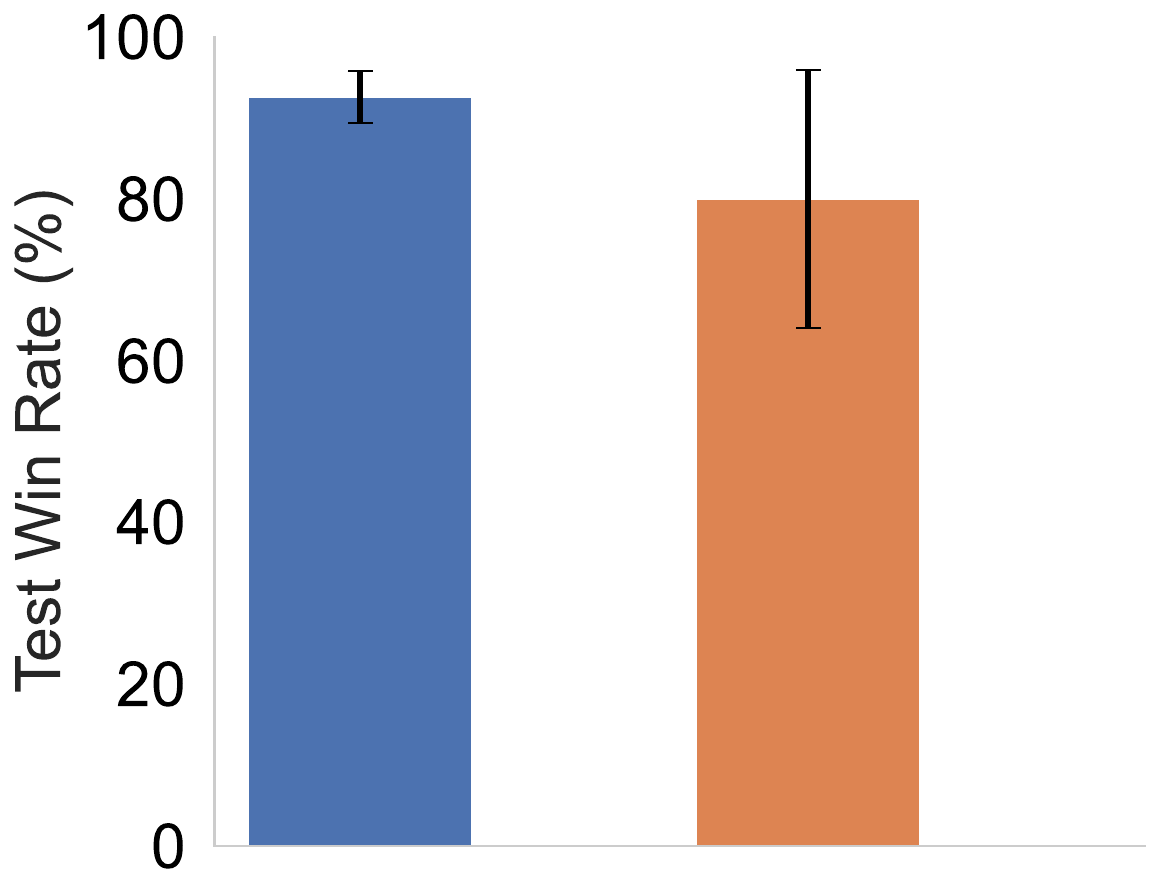}}
    
   \vspace{0.36em}    
    \caption{Ablation study on residual connection.} \label{residual}
 \vspace{-0.3cm}
\end{figure}

\begin{figure*}[!t]
    \centering
    \subfloat{\quad\quad\includegraphics[width=0.6\textwidth]{fig/legend1.pdf}}\\    
    \addtocounter{subfigure}{-1}
    \vspace{-0.1cm}
    \subfloat[protoss\_5\_5]{\includegraphics[width=0.33\textwidth]{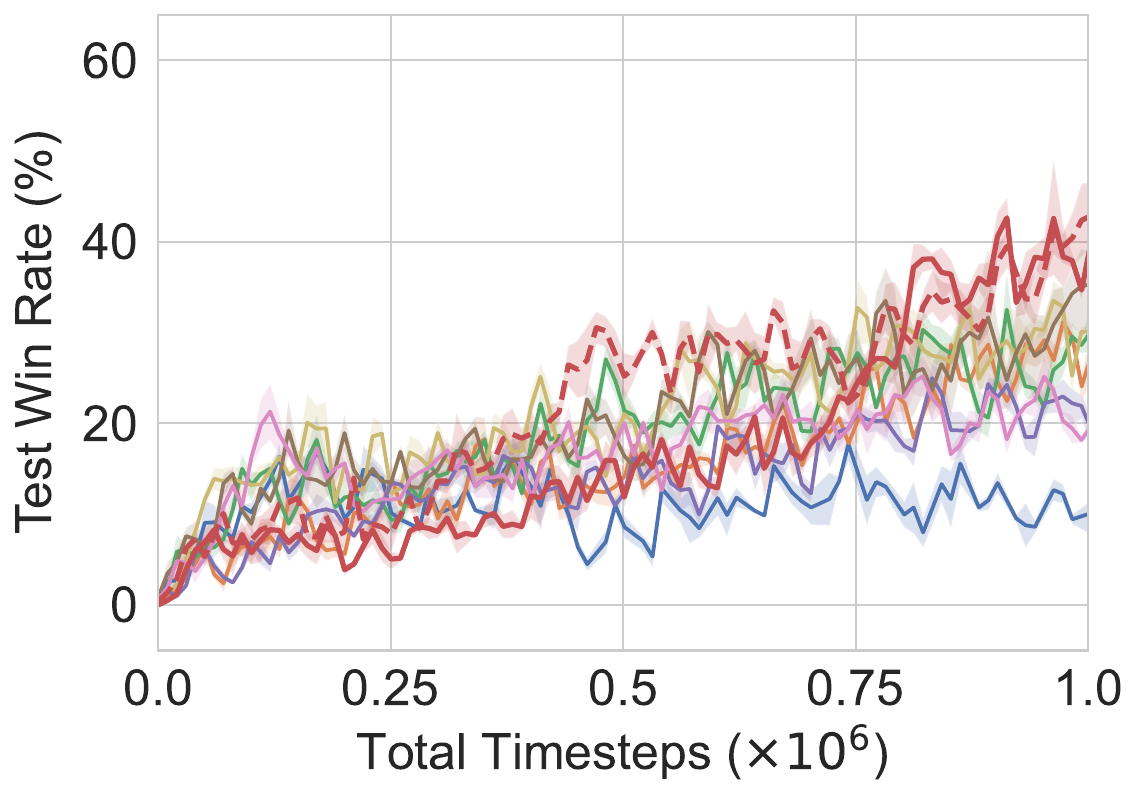}}
    \subfloat[terran\_5\_5]{\includegraphics[width=0.33\textwidth]{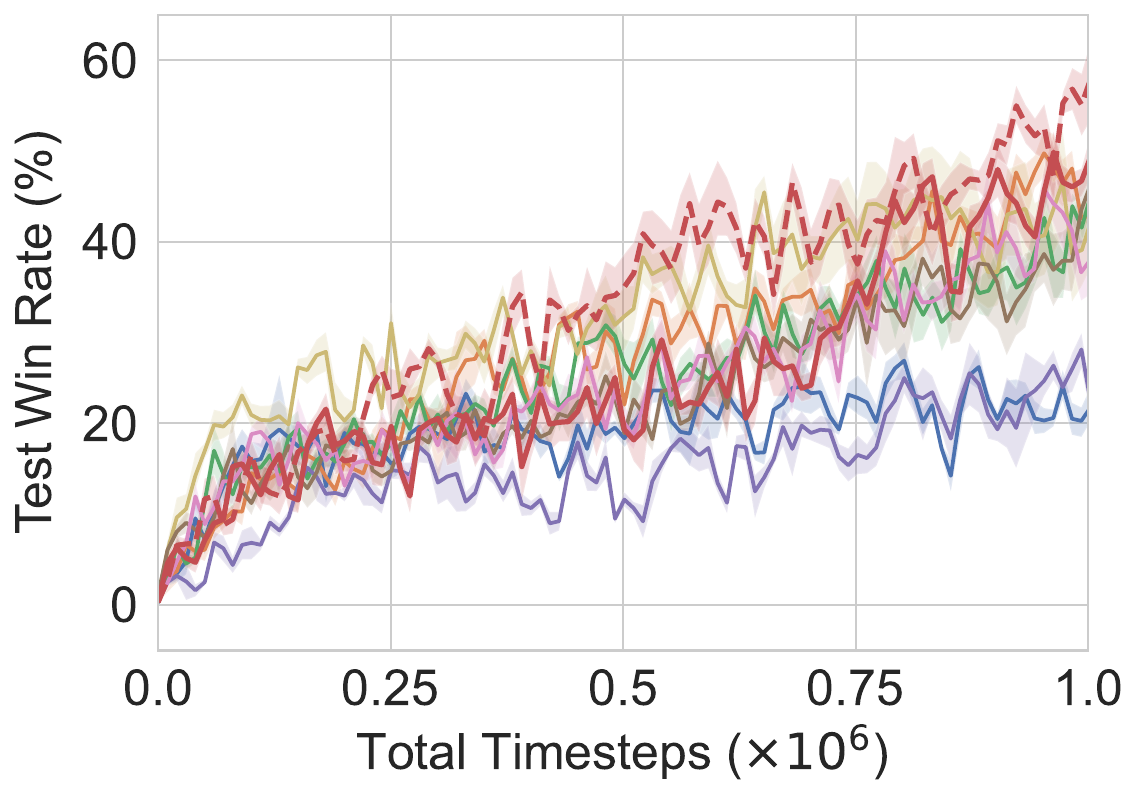}}
    \subfloat[zerg\_5\_5]{\includegraphics[width=0.33\textwidth]{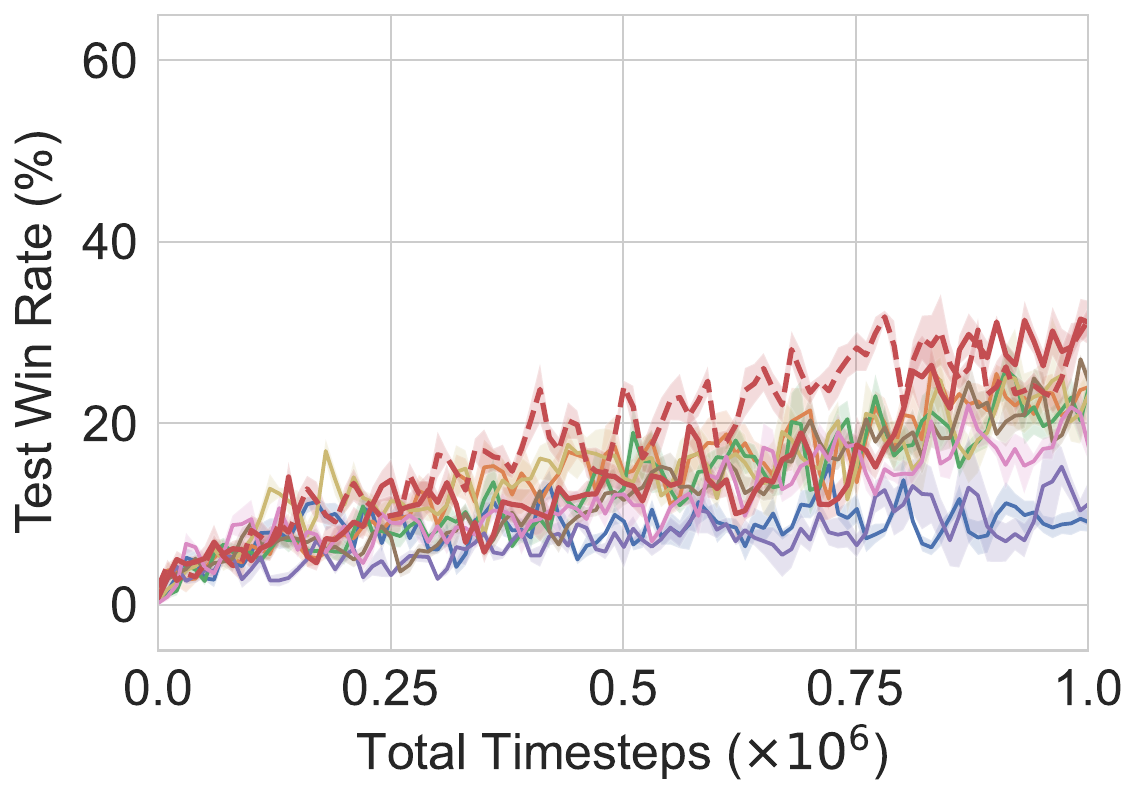}} \\
    \caption{Learning curves of our method and baselines on the SMACv2 scenarios. }
    \label{smacv2}
  \end{figure*}


\begin{figure}[!t]
    \centering
     \subfloat{\includegraphics[width=0.33\textwidth]{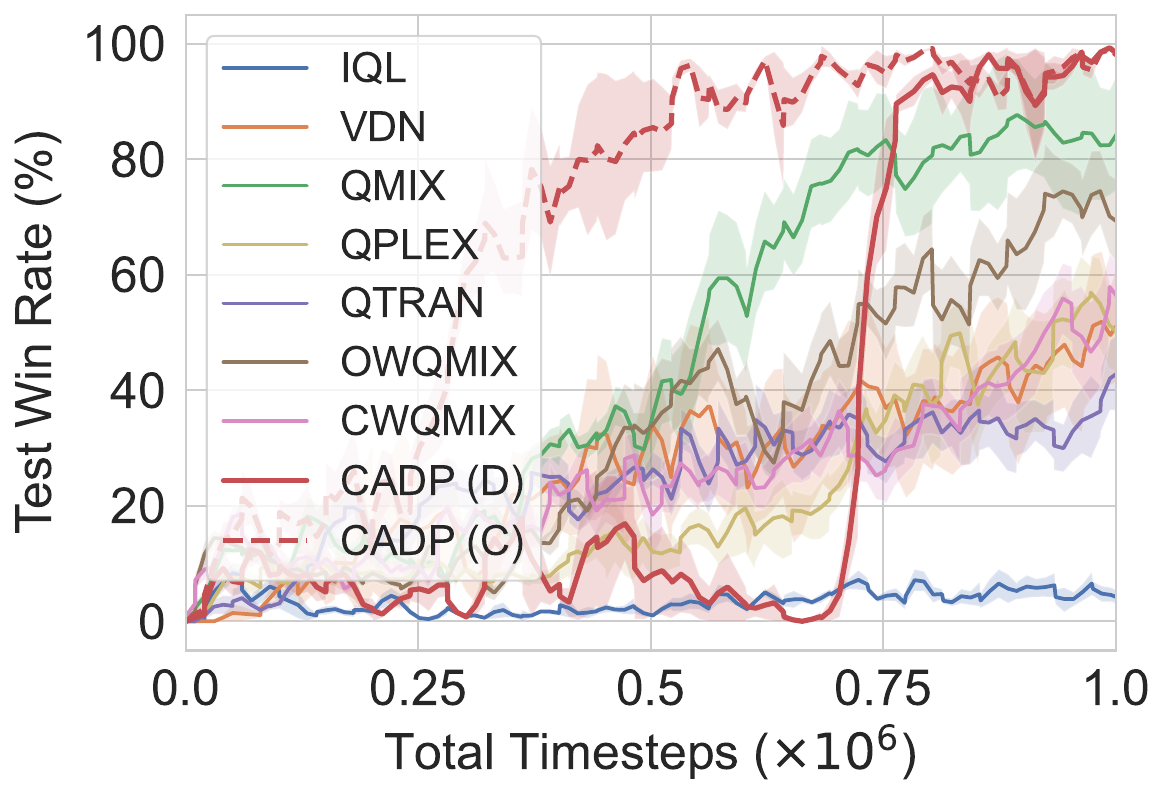}}
    \caption{Learning curves on  the 25m scenario.}
    \label{fig:25m}
  \end{figure}

\subsection{Detailed Results of SMAC and GRF}
\textbf{SMAC.}
The StarCraft Multi-Agent Challenge~(SMAC)\footnote{We use SC2.4.10 version instead of the older SC2.4.6.2.69232. Performance is not comparable across versions.}~\cite{SMAC} has become a common-used benchmark for evaluating state-of-the-art MARL methods. SMAC focuses on micromanagement challenges where each of the ally entities is controlled by an individual learning agent, and enemy entities are controlled by a built-in AI. 
At each time step, agents can move in four cardinal directions, stop, take no-operation, or choose an enemy to attack. Thus, if there are $N_e$ enemies in the scenario, the action space for each ally unit consists of $N_e + 6$ discrete actions. Figure~\ref{fig:sc2} shows the initial state of three SMAC scenarios.
The goal of the agents is to maximize the damage to enemies. Hence, proper tactics such as sneaking attack and drawing fire are required during battles. Learning these diverse interaction behaviors under partial observation is a challenging task.
To validate the effectiveness of our methods, we conduct experiments on 3 SMAC scenarios~\cite{SMAC} which are classified into \textbf{Hard}~(\emph{5m\_vs\_6m}) and \textbf{Super Hard}~(\emph{corridor}, \emph{3s5z\_vs\_3s6z}), since many easy SMAC scenarios already can be solved well by existing basic CTDE methods.

\textbf{GRF.}
Google Research Football~(GRF)~\cite{GRF} is another MARL benchmark based on open-source football game.  Figure~\ref{fig:grf_picture} shows two GRF scenarios.
In our experiments, we control left-side players except the goalkeeper. The right-side players are rule-based
bots controlled by the game engine. Agents have a discrete action space of 19, including moving in
8 directions, sliding, shooting, and passing. The observation contains the positions and moving
directions of the ego-agent, other agents, and the ball. The z-coordinate of the ball is also included. We test our approach on two challenging GRF offensive scenarios: \emph{academy\_3\_vs\_1\_with\_keeper}~(three of our players try to score from the edge of the box, one on each side, and the other at the center. Initially, the player at the center has the ball and is facing the defender. There is an opponent keeper) and  \emph{academy\_counterattack\_easy}~(4 versus 1 counter-attack with keeper; all the remaining players of both teams run back towards the ball). Unlike StarCraft II, individual observations in the Google Research Football~(GRF)  are not partial observation, which contains as much information as global state, but the global state uses absolute coordinates, while personal observations use relative coordinates based on the own positions of agents.  

 Please refer to the main manuscript for the learning curves of compared methods. Here we provide another quantitative metric to compare different methods. We calculate the average test win rate over the last 100K training steps for each model due to the unstable fluctuation of performance. The additional quantitative results are reported in Table~\ref{tab:result_sc2} and Table~\ref{tab:result_GRF}. Our method outperforms various baselines by a large margin.

 \algdef{SE}[SUBALG]{Indent}{EndIndent}{}{\algorithmicend\ }%
\algtext*{Indent}
\algtext*{EndIndent}
\algnewcommand{\LineComment}[1]{\State \textcolor{gray}{\(\#\) #1}}

\begin{algorithm}[!t]
  \caption{Centralized Training in CADP}
  \label{alg:alg1}
  \begin{algorithmic}[1]
    \Statex \textbf{Initialize:} agent network $\theta_{a}$,\,
           mixing network $\theta_{\upsilon}$,\,
           target network $\theta_{\upsilon}^-=\theta_{\upsilon}$,\,
           replay buffer $\mathcal{D}$
    \Repeat
      \LineComment{Collect the data}
      \State Obtain the initial global state $s_0$
      \While{not terminal}
        \For{each agent $i$}
          \State Obtain the observation $o_i^t = O(s_t,i)$
          \LineComment{Message Exchanging}
          \State Calculate $\{q_i^t,k_i^t,v_i^t\}$
          \State Send $\{k_i^t,v_i^t\}$ to other agents
          \State Receive $\{k_{1:N}^t,v_{1:N}^t\}$ from others
          \State $z_i^t = \sum_{j=1}^{N} c_{i,j}\,v_j^t$
          \LineComment{Policy}
          \State $h_i^t = \operatorname{GRU}([z_i^t,o_i^t],h_{i}^{t-1})$
          \State Compute $Q_t^i(h_i^t,\cdot)$
          \State Sample $u_t^i$ from $Q_t^i(h_i^t,\cdot)$ by $\epsilon$-greedy
        \EndFor
        \State Execute the joint action $\boldsymbol{u}_t=[u_t^1,\dots,u_t^N]$
        \State Receive reward $r_{t+1}$ and next state $s_{t+1}$
      \EndWhile
      \State Store the episode in buffer $\mathcal{D}$
      \LineComment{Train the networks}
      \State Sample episodes from $\mathcal{D}$
      \State Calculate joint action values with mixing network
      \State Compute $\mathcal{L}_{TD}(\theta_{mix},\theta_a)$, $\mathcal{L}_{p}(\theta_a)$
      \State Update $\theta_{mix}$ and $\theta_{a}$ by gradient descent
      \State Update target networks $\theta_{mix}^-=\theta_{mix}$,\,
             $\theta_{a}^-=\theta_{a}$ every $C$ episodes
    \Until{converge}
  \end{algorithmic}
\end{algorithm}

\begin{algorithm}[!t]
  \caption{Decentralized Execution in CADP}
  \label{alg:alg2}
  \begin{algorithmic}[1]
    \Statex \textbf{Input:} agent network $\theta_{a}$
    \State Obtain the initial global state $s_0$
    \While{not terminal}
      \For{each agent $i$}
        \State Obtain the observation $o_i^t = O(s_t,i)$
        \LineComment{without Message Exchanging}
        \State Calculate $v_i^t$
        \LineComment{Policy}
        \State $h_i^t = \operatorname{GRU}([v_i^t,o_i^t],h_{i}^{t-1})$
        \State Compute $Q_t^i(h_i^t,\cdot)$
        \State $u_t^i = \arg\max Q_t^i(h_i^t,\cdot)$
      \EndFor
      \State Execute the joint action $\boldsymbol{u}_t=[u_t^1,\dots,u_t^N]$
      \State Receive reward $r_{t+1}$ and next state $s_{t+1}$
    \EndWhile
  \end{algorithmic}
\end{algorithm}
  
\begin{table}[!t]
    \centering
    \small
    \caption{The test win rate of compared methods on the SMAC scenarios~(\emph{5m\_vs\_6m}, \emph{corridor} and \emph{3s5z\_vs\_3s6z}). $\pm$ corresponds to one standard deviation of the average evaluation over 5 trials.}
    \vspace{0.5em}
    \label{tab:result_sc2}
     \fontsize{8pt}{8pt}\selectfont 
    \begin{tabular}{@{}lcccccc@{}}
    \toprule
     \multicolumn{1}{c}{\textbf{Method}} & \textbf{5m\_vs\_6m} & \textbf{corridor}    & \textbf{3s5z\_vs\_3s6z}  \\ \midrule
    \textbf{IQL} & 0.13 $\pm$ 0.06  & 0.50 $\pm$ 0.15  & 0.00 $\pm$ 0.00  \\ \specialrule{0em}{1pt}{1pt}
    \textbf{VDN}  & 0.54 $\pm$ 0.09  & 0.65 $\pm$ 0.32  & 0.25 $\pm$ 0.18  \\ \specialrule{0em}{1pt}{1pt}
    \textbf{QMIX}  & 0.43 $\pm$ 0.13  & 0.70 $\pm$ 0.35  & 0.24 $\pm$ 0.36  \\ \specialrule{0em}{1pt}{1pt}
    \textbf{QPLEX}  & 0.57 $\pm$ 0.13   & 0.20 $\pm$ 0.12  & 0.08 $\pm$ 0.11  \\ \specialrule{0em}{1pt}{1pt}
    \textbf{QTRAN}  & 0.28 $\pm$ 0.08   & 0.00 $\pm$ 0.00  & 0.00 $\pm$ 0.00  \\ \specialrule{0em}{1pt}{1pt}
    \textbf{OWQMIX}  & 0.52 $\pm$ 0.08  & 0.16 $\pm$ 0.33  & 0.30 $\pm$ 0.23  \\ \specialrule{0em}{1pt}{1pt}
    \textbf{CWQMIX}  & 0.44 $\pm$ 0.15  & 0.00 $\pm$ 0.00  & 0.19 $\pm$ 0.16  \\ \specialrule{0em}{1pt}{1pt} 
    \textbf{IGM-DA~(Teacher)}  & 0.31 $\pm$ 0.07  & 0.00 $\pm$ 0.00  & 0.06 $\pm$ 0.09  \\ \specialrule{0em}{1pt}{1pt} 
    \textbf{IGM-DA~(Student)}  & 0.54 $\pm$ 0.09  & 0.00 $\pm$ 0.00  & 0.18 $\pm$ 0.22  \\ \specialrule{0em}{1pt}{1pt} 
    \textbf{CTDS~(Teacher)}  & 0.55 $\pm$ 0.13  & 0.36 $\pm$ 0.44  & 0.69 $\pm$ 0.35  \\ \specialrule{0em}{1pt}{1pt} 
    \textbf{CTDS~(Student)}  & 0.51 $\pm$ 0.13  & 0.34 $\pm$ 0.42  & 0.57 $\pm$ 0.30  \\ \specialrule{0em}{1pt}{1pt} 
    \hline \specialrule{0em}{1pt}{1pt} 
    \textbf{CADP~(C)}  & \textbf{0.68 $\pm$ 0.08}  & \textbf{0.85 $\pm$ 0.04}  & \textbf{0.94 $\pm$ 0.03}  \\ \specialrule{0em}{1pt}{1pt} 
    \textbf{CADP~(D)}  & \textbf{0.68 $\pm$ 0.08}  & \textbf{0.84 $\pm$ 0.03}  & \textbf{0.93 $\pm$ 0.03}  \\ \specialrule{0em}{1pt}{1pt} \bottomrule
    \end{tabular}%
    
    \end{table}

    \begin{table}[!t]
    \centering
    \small
    \caption{The test win rate of compared methods on the GRF scenarios~(\emph{3\_vs\_1\_ with\_keeper} and \emph{counterattack\_easy}). $\pm$ corresponds to one standard deviation of the average evaluation over 5 trials.}
    \label{tab:result_GRF}
    \vspace{0.5em}
 
    \fontsize{8pt}{8pt}\selectfont 
    \begin{tabular}{@{}lcccc@{}}
    \toprule
     \multicolumn{1}{c}{\textbf{Method}}& \textbf{3\_vs\_1\_with\ \_keeper} & \textbf{counterattack\_easy}  \\ \midrule
    \textbf{QMIX}  & 0.58 $\pm$ 0.21  & 0.24 $\pm$ 0.13  \\ \specialrule{0em}{1pt}{1pt}
    \textbf{IGM-DA~(Teacher)}  & 0.05 $\pm$ 0.03  & 0.11 $\pm$ 0.08 \\ \specialrule{0em}{1pt}{1pt} 
    \textbf{IGM-DA~(Student)}  & 0.05 $\pm$ 0.02  & 0.16 $\pm$ 0.14 \\ \specialrule{0em}{1pt}{1pt} 
    \textbf{CTDS~(Teacher)}  & 0.61 $\pm$ 0.20  & 0.38 $\pm$ 0.29 \\ \specialrule{0em}{1pt}{1pt} 
    \textbf{CTDS~(Student)}  & 0.58 $\pm$ 0.22  & 0.33 $\pm$ 0.27 \\ \specialrule{0em}{1pt}{1pt} 
    \hline \specialrule{0em}{1pt}{1pt} 
    \textbf{CADP~(C)}  & \textbf{0.77 $\pm$ 0.00}  & \textbf{0.64 $\pm$ 0.15}  \\ \specialrule{0em}{1pt}{1pt} 
    \textbf{CADP~(D)}  & \textbf{0.79 $\pm$ 0.01}  & \textbf{0.56 $\pm$ 0.18}  \\ \specialrule{0em}{1pt}{1pt} \bottomrule
    \end{tabular}%
    
    \end{table}

\subsection{Different sight range.}
We  test  methods in the complex \emph{3s5z\_vs \_3s6z} scenario with different sight range of 9, 5, and 0. Sight range represents the radius of field of view for each agent in the game. Figure~\ref{fig:sightrange} shows the experimental results.
We can see that as the sight range  becomes smaller, the performance of most methods  decline since smaller sight range means less information will be collected by agents. But the performance of QMIX and IGM-DA with sight range 5 is higher than with sight range 9, which may be because in some situations,  "Limiting the value function input dimensionality can
further improve performance"~\cite{MAPPO}. Anyway, our method still has the best performance with all different sight ranges and our method still is the only one which can ultimately reach almost consistent performance for decentralized model and centralized model in all three settings.  Since there may be an  inevitable information gap between teacher with global information and student with partial observation especially when the sight range is very small. Then teacher may teach beyond the student's ability. But our pruning loss enables our centralized mode to adjust strategy based on the abilities of decentralized model instead of blindly imitation learning.

\subsection{Residual connection.}
To validate the effect of residual connection, we conduct ablation experiments by removing and adding the residual connection module. The experimental results shown in Figure~\ref{residual}  demonstrate that adding the residual connection module significantly enhances the stability and performance of the model.

\subsection{Additional Results for SMACv2}
Since recently SMAC was preceived as lacking stochastic partial observability and the upgraded version of the SMAC environment, SMACv2, was  proposed and considered a better MARL benchmark~\cite{messy_smac,ellis2022smacv2}, we also conducted experiments on SMACv2. For fairness and comparability, we directly follow the parameters settings of the original SMAC version to conduct experiments on SMACv2 for all methods.  As shown in Figure~\ref{smacv2}, the performance of our method remains superior to all baselines.

\begin{figure*}[!b]
    \centering
    \subfloat[The heat map of the attention matrix]{\quad\quad\includegraphics[width=0.92\textwidth]{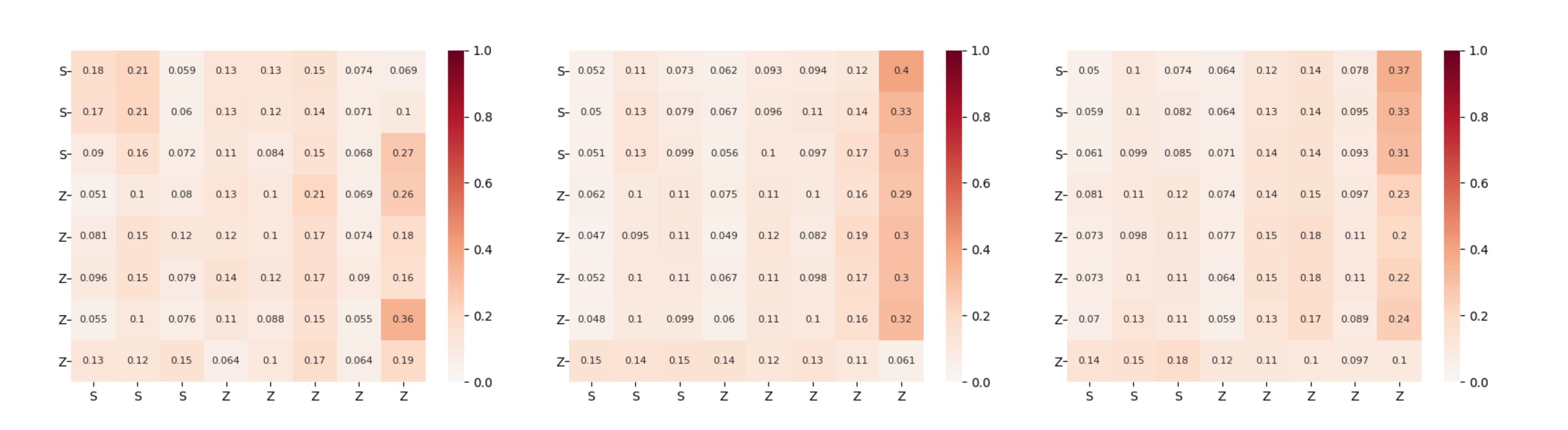}}\\
   \subfloat[The behavior of the centralized model]{\includegraphics[width=0.80\textwidth]{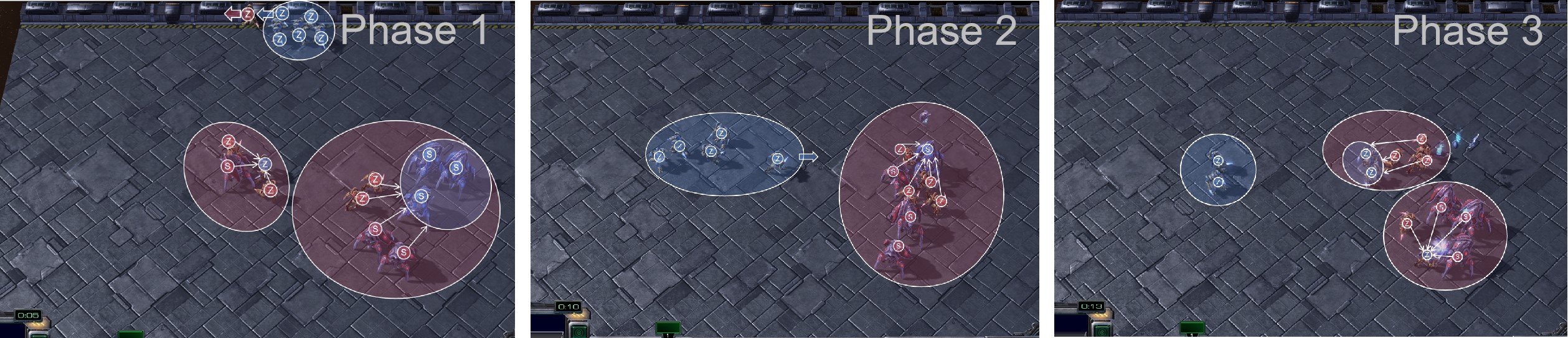}}\\
    \subfloat[The behavior of the decentralized model]{\includegraphics[width=0.80\textwidth]{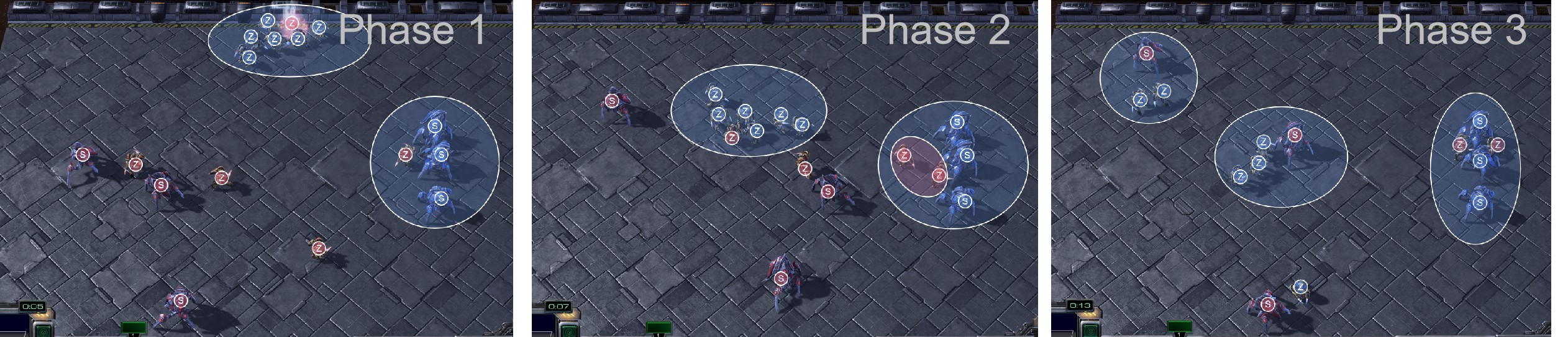}}
    \vspace{-0.5em}
    \caption{ A visualization example of the attack strategies that emerge at timestep 3M. The top is the heat map of the attention matrix, the middle is the behavior of the centralized model and the bottom is the behavior of the decentralized model. Red and blue shadows indicate the agent and enemy formations, respectively. Red and blue arrows indicate the moving direction of the agent and enemy formations, respectively. White arrows indicate  attack action from the agent units to the enemy units.}
    \label{fig:visual0}
  \end{figure*}

\subsection{Additional Results for Large Number of Agents}
We  conduct experiments in the 25m scenario with 25 controllable agents. The 25m scenario is one of the scenarios with the highest number of controllable agents in the SMAC~\cite{SMAC} environment. As shown in Figure~\ref{fig:25m}, the results demonstrate that our proposed CADP framework can learn effective attention weights and offer promising results with tens of agents.

\subsection{Comparison of Computational Time}\label{time_sec}

\begin{table}[!b]
    \centering
    \small
    \caption{The runtimes of different methods in the 5m\_vs\_6m scenario for 1M timesteps.}
    \label{time}

  \begin{tabular}{cccc} 
    \toprule
    \multicolumn{1}{c}{\textbf{Method}} & \textbf{Total (Hour)} & \textbf{Sample (Hour)} & \textbf{Train (Hour)} \\
    \midrule
    QMIX & 8.17 $\pm$ 0.56 & 5.63 $\pm$ 0.06 & 1.90 $\pm$ 0.8\;\, \\ \specialrule{0em}{1pt}{1pt}
    OWQMIX & 9.25 $\pm$ 0.58 & 5.63 $\pm$ 0.13 & 2.96 $\pm$ 0.58 \\ \specialrule{0em}{1pt}{1pt}
    QPLEX & 10.13 $\pm$ 0.13 & 5.83 $\pm$ 0.02 & 3.71 $\pm$ 0.08 \\ \specialrule{0em}{1pt}{1pt}\midrule
    IGM-DA & 9.43 $\pm$ 0.33 & 5.72 $\pm$ 0.03 & 3.12 $\pm$ 0.13 \\ \specialrule{0em}{1pt}{1pt}
    CTDS & 9.55 $\pm$ 0.33 & 5.66 $\pm$ 0.21 & 3.34 $\pm$ 0.21 \\ \specialrule{0em}{1pt}{1pt}
    \midrule
    CADP & {10.50 $\pm$ 0.28} & {6.17 $\pm$ 0.06} & {3.58 $\pm$ 0.97} \\
    \bottomrule
    
  \end{tabular}
\end{table}%
We  record the runtimes of three types in the 5m\_vs\_6m scenario for 1 million time-steps: total runtime, runtime for sampling only, and runtime for training only. 
As shown in Table \ref{time}, the results reveal that the runtime for different methods mainly comes from the sampling process, which involves interacting with the environment. CADP only incurs a small additional overhead due to communication.

\section{Pseudocode of CADP}\label{code}
To make the proposed method clearer, the pseudocode is provided in Algorithm~\ref{alg:alg1} and Algorithm~\ref{alg:alg2}. Here we take the method of value decomposition as an example. For the policy gradient (PG) method, we just need to add the pruning loss to the loss of actor.

\section{Visualization} \label{visual}
\begin{figure*}[!h]
    \centering
    \subfloat[The heat map of the attention matrix]{\quad\quad\includegraphics[width=0.92\textwidth]{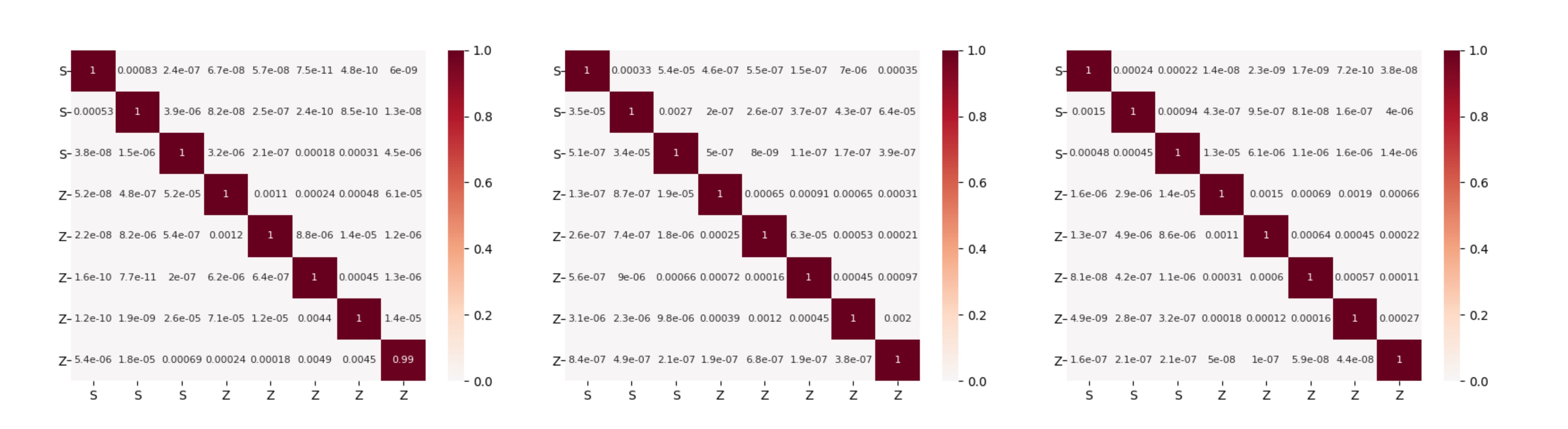}}\\
    \subfloat[The behavior of the centralized model]{\includegraphics[width=0.80\textwidth]{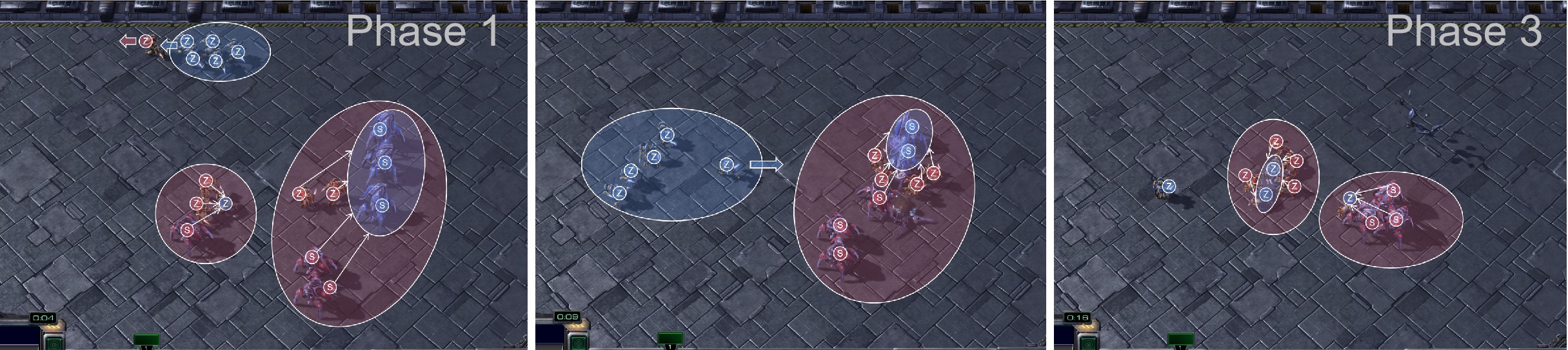}}\\
    \subfloat[The behavior of the decentralized model]{\includegraphics[width=0.80\textwidth]{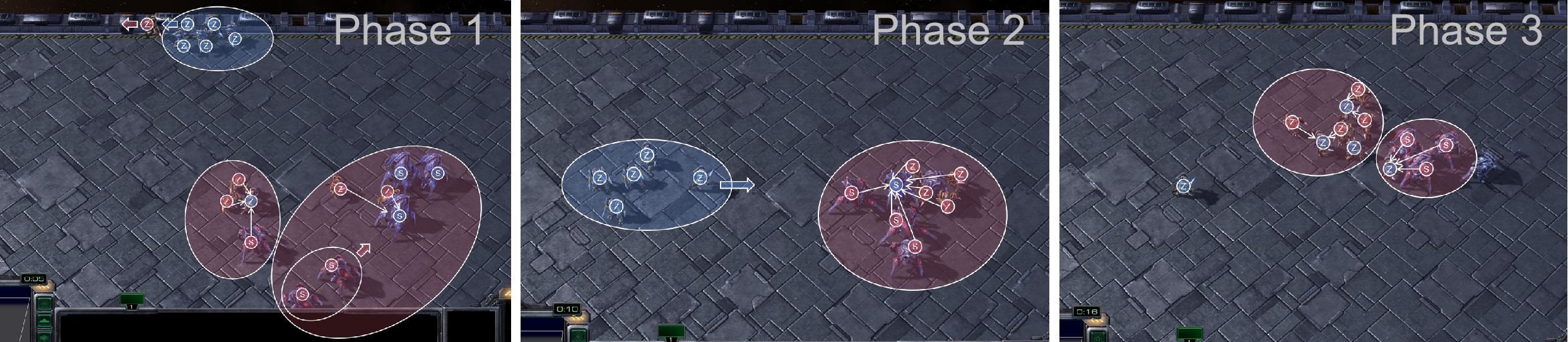}}
    \vspace{-0.5em}
    \caption{ A visualization example of the attack strategies that emerge in the \emph{3s5z\_vs\_3s6z} scenario at timestep 3.5M, after adding  pruning loss $\mathcal{L}_p$ for 0.5M timesteps. }
    \label{fig:visual1}
  \end{figure*}

To further explain the working process of the pruning loss function  $ \mathcal{L}_{p}$, we designed a set of qualitative analysis in Figure~\ref{fig:visual0} and ~\ref{fig:visual1}.

We can find in \emph{3s5z\_vs\_3s6z} the centralized model quickly weans itself off the  advising and the decentralized model learns the similar strategy to the centralized model within 0.5M timesteps. 
At timestep 3M, the centralized model already learns team-oriented strategy. To be specific, firstly deploy a Zealot to distract the enemy's main force~(5 Zealots). This zealot makes the enemy units give chase and maintain enough distance so that little  damage is incurred while our main force splits the enemy into two parts and wipes them out in groups. After the enemy's 5 Zealots kill our victim unit~(a Zealot) and turn back, our force already occupies the numerical advantage. Then our force is divided into two groups, focusing fire on the enemy's remaining troops. However, the decentralized model at the same timestep does not have a good strategy for actions. It shows that the victim unit does not work well as it be killed by the enemy's main force quickly, meanwhile, our main force does not organize an effective attack. So that when the enemy's main force turn back, they occupy the numerical advantage and soon they wipe out our army. We can see that although the decentralized model has some similarities in their action strategy with centralized model since the decentralized model is included in the centralized model, but the decentralized model does not work well as the pruning loss function  $ \mathcal{L}_{p}$  still not be added and the model is now highly depended on suggestions of other agents as shown in  Figure~\ref{fig:visual0}(a). 

Then after adding the $ \mathcal{L}_{p}$ for 0.5M timesteps, we can  see that the model is now almost no need for the suggestions of other agents as shown in  Figure~\ref{fig:visual1}(a).  The centralized model and the decentralized model at timesetp 3.5M almost execute highly consistent strategy which is also similar to the strategy of the centralized model at timestep 3M. In a word, by adding the pruning loss function  $ \mathcal{L}_{p}$, the decentralized model quickly learns the team-orient strategy of centralized model.


\end{document}